\documentclass{article}
\usepackage{arxiv}

\usepackage[utf8]{inputenc} 
\usepackage[T1]{fontenc}    
\usepackage{hyperref}       
\usepackage{url}            
\usepackage{booktabs}       
\usepackage{amsfonts}       
\usepackage{nicefrac}       
\usepackage{microtype}      
\usepackage{lipsum}
\usepackage{graphicx}
\graphicspath{ {./images/} }

\usepackage{amsmath,amssymb,amsthm}
\usepackage{float}
\usepackage[caption = false]{subfig}

\usepackage[square,numbers,sort&compress]{natbib}
\usepackage[caption = false]{subfig}
\usepackage[ruled,vlined]{algorithm2e}
\usepackage{comment}
\newcommand{\f}{\mathbf{f}}
\newcommand{\g}{\mathbf{g}}
\newcommand{\w}{\mathbf{w}}
\newcommand{\m}{\mathbf{m}}
\newcommand{\K}{\mathbf{K}}
\newcommand{\cov}{\mathrm{cov}}
\newcommand{\kl}{\mathrm{KL}}
\newcommand{\var}{\mathrm{var}}
\newcommand{\R}{\mathbb{R}}

\newcommand{\E}{\mathbb{E}}
\newcommand{\A}{\mathrm{A}}

\newcommand{\dt}{\Delta_{\boldsymbol{\theta}}}
\newcommand{\bt}{\boldsymbol{\theta}}
\newcommand{\dtone}{\Delta_{\boldsymbol{\theta}^1}}
\newcommand{\dttwo}{\Delta_{\boldsymbol{\theta}^2}}

\newcommand{\thetaobs}{\theta_{\mathrm{obs}}}
\newcommand{\btobs}{\boldsymbol{\theta}_{\mathrm{obs}}}

\newcommand{\x}{\mathbf{x}}
\newcommand{\xobs}{\mathbf{x}_{\mathrm{obs}}}
\newcommand{\xtheta}{\mathbf{x}_{\theta}}

\newcommand{\X}{\mathbf{X}}
\newcommand{\Xobs}{\mathbf{X}_{\mathrm{obs}}}
\newcommand{\Xtheta}{\mathbf{X}_{\theta}}

\newcommand{\Xthetafirst}{\mathbf{X}_{\theta^1}}
\newcommand{\Xthetasecond}{\mathbf{X}_{\theta^2}}
\newcommand{\sthetafirst}{\mathbf{s}_{\theta^1}}
\newcommand{\sthetasecond}{\mathbf{s}_{\theta^2}}

\newcommand{\s}{s}
\newcommand{\sobs}{\mathbf{s}_{\mathrm{obs}}}
\newcommand{\stheta}{\mathbf{s}_{\theta}}

\newcommand{\W}{W_{D}^*}
\newcommand{\Y}{\mathbf{Y}}
\newcommand{\y}{\mathbf{y}}
\newcommand\eatpunct[1]{}

\begin{document}

\title{Likelihood-Free Inference with Deep Gaussian Processes}

\author{Alexander Aushev  \\
  Helsinki Institute for Information Technology, \\
  Department of Computer Science, \\
  Aalto University, Finland \\
  \texttt{alexander.aushev@aalto.fi} \\
  \And
  Henri Pesonen \\
  Department of Biostatistics, \\
  University of Oslo, Norway \\
  \texttt{henri.pesonen@medisin.uio.no} \\
  \And
  Markus Heinonen \\
  Helsinki Institute for Information Technology, \\
  Department of Computer Science, \\
  Aalto University, Finland \\
  \texttt{markus.o.heinonen@aalto.fi} \\
  \And
  Jukka Corander \\
  Department of Biostatistics, \\
  University of Oslo, Norway \\
  Department of Mathematics and Statistics, \\
  University of Helsinki, Finland \\
  \texttt{jukka.corander@medisin.uio.no}
  \And
  Samuel Kaski \\
  Helsinki Institute for Information Technology, \\
  Department of Computer Science,\\
  Aalto University, Finland \\
  Department of Computer Science, \\
  University of Manchester, UK \\
  \texttt{samuel.kaski@aalto.fi} \\
}

\maketitle

\begin{abstract}
In recent years, surrogate models have been successfully used in likelihood-free inference to decrease the number of simulator evaluations. The current state-of-the-art performance for this task has been achieved by Bayesian Optimisation with Gaussian Processes (GPs). While this combination works well for unimodal target distributions, it is restricting the flexibility and applicability of Bayesian Optimisation for accelerating likelihood-free inference more generally. We address this problem by proposing a Deep Gaussian Process (DGP) surrogate model that can handle more irregularly behaved target distributions. Our experiments show how DGPs can outperform GPs on objective functions with multimodal distributions and maintain a comparable performance in unimodal cases. At the same time, DGPs  generally require much fewer data to achieve the same level of performance as neural density and kernel mean embedding alternatives. This confirms that DGPs as surrogate models can extend the applicability of Bayesian Optimisation for likelihood-free inference (BOLFI), while only adding computational overhead that remains negligible for computationally intensive simulators. 
\end{abstract}

\section{Introduction}
\label{sec:1-intro}

Likelihood-free inference (LFI) for simulator-based models has been a topic of substantial interest during the past two decades for the computational modelling community \citep{hartig2011, elfi2018}. In LFI, we aim to infer the generative parameters $\bt$ of an observed dataset $\Xobs = \{ \xobs \}$, whose likelihood $p(\xobs | \bt)$ is intractable, which prevents the conventional statistical parameter estimation \citep{diggle1984}. Instead, we assume we can simulate new data $\{\xtheta\} \sim p(\x|\bt)$ using any feasible parameter values. We relate the probability of a parameter to how similar its simulated dataset $\Xtheta$ is to the observed one \citep{hartig2011}, measured via a discrepancy function. Different simulator-based LFI approaches have been proposed under the names of approximate Bayesian computation (ABC) \citep{beaumont2002, sunnaaker2013approximate, csillery2010approximate, beaumont2009adaptive}, indirect inference \citep{gourieroux2010indirect, genton2003robust, heggland2004estimating} and synthetic likelihood \citep{wood2010, price2018bayesian, ong2018variational} in domains ranging from genetics \citep{beaumont2002, pritchard1999population, boitard2016inferring} to economics \citep{guvenen2010inferring, monfardini1998estimating} and ecology \citep{wood2010, beaumont2010approximate, van2015calibration}.

When the simulator call-time is long, the number of simulator queries has to be limited for computational reasons. Therefore, a popular trend in LFI literature combines traditional methods with active learning \citep{settles2009active, rubens2015active} to improve sample-efficiency. For instance, some neural-based density estimations \citep{alsing2019fast, lueckmann2019likelihood, greenberg2019automatic, papamakarios2019} and kernel mean embedding methods \citep{chowdhury2020active, hsu2019bayesian} allow high-fidelity posterior inference with thousands of samples. A particularly suitable approach to LFI in this setting is to find a data-efficient surrogate to the discrepancy function, which can be used to derive a proxy for the unknown likelihood. Previous research by \citet{gutmann2016bayesian} has addressed this issue by using Gaussian Processes (GPs) as the discrepancy surrogates and applying Bayesian Optimisation (BO) as an efficient search strategy. This approach drastically reduced the number of simulations required for accurate inference, to the order of only hundreds. 

\begin{figure*}
    \centering
    \hfill
    \subfloat[True $\dt$]{\label{fig:1-true}\includegraphics[height=3.9cm]{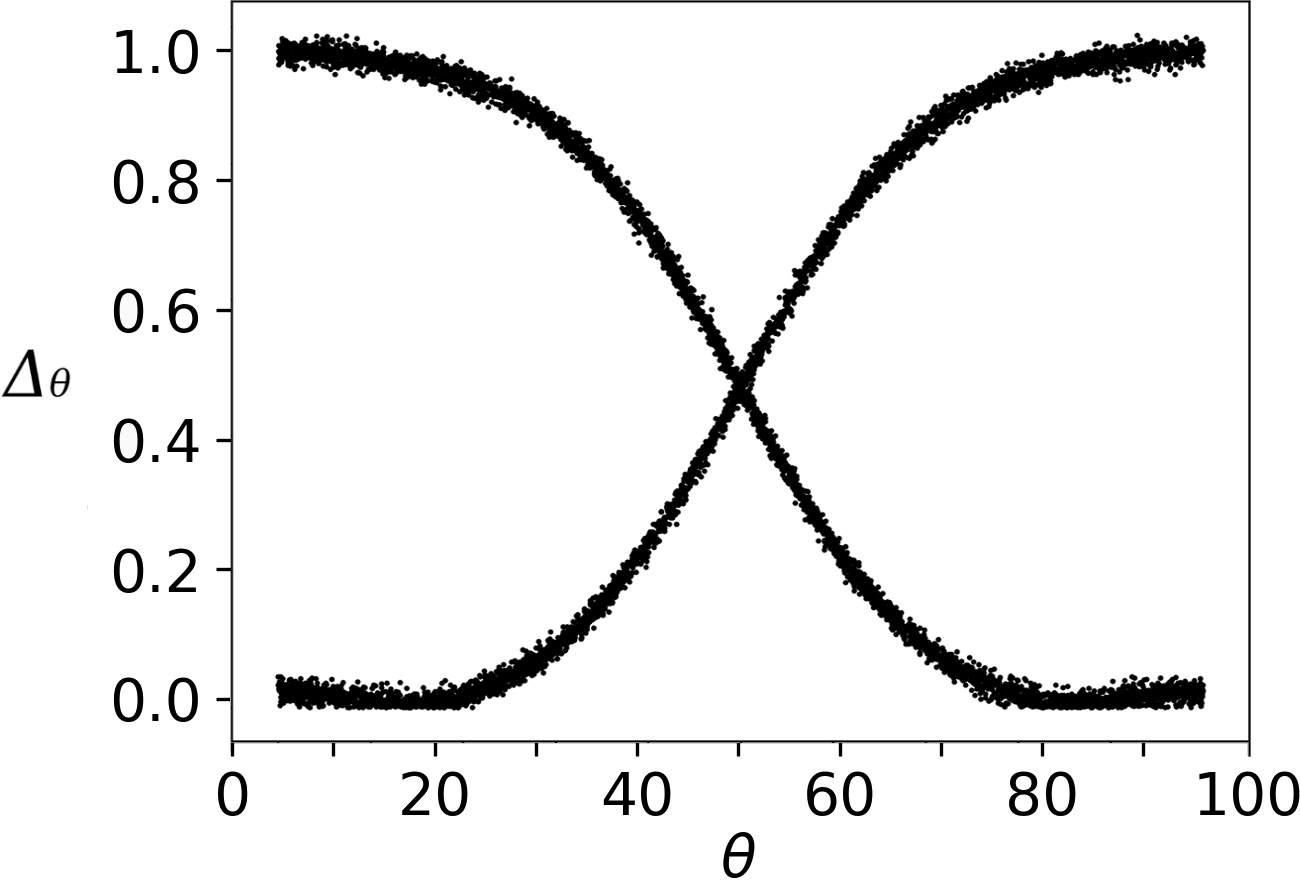}}
    \hfill
    \subfloat[Vanilla GP fit]{\label{fig:1-gp}\includegraphics[height=3.9cm]{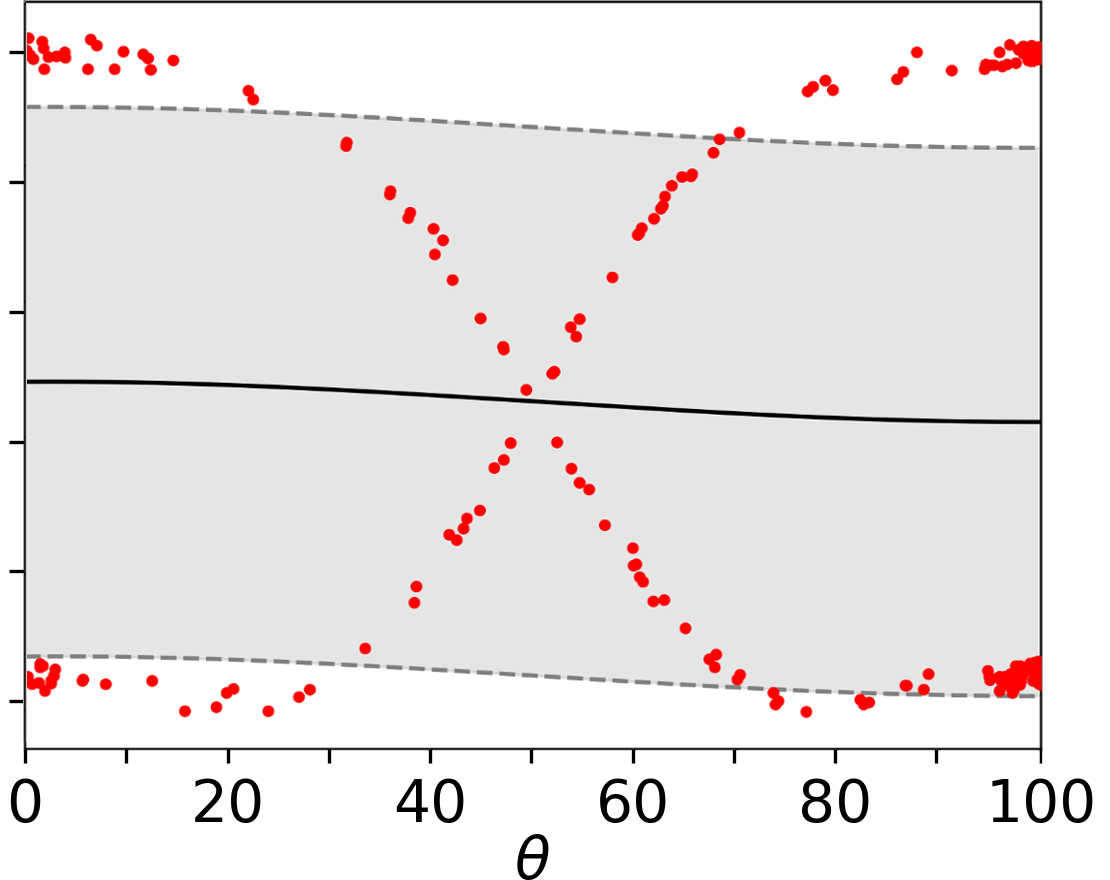}}
    \hfill
    \subfloat[DGP fit]{\label{fig:1-dgp}\includegraphics[height=3.9cm]{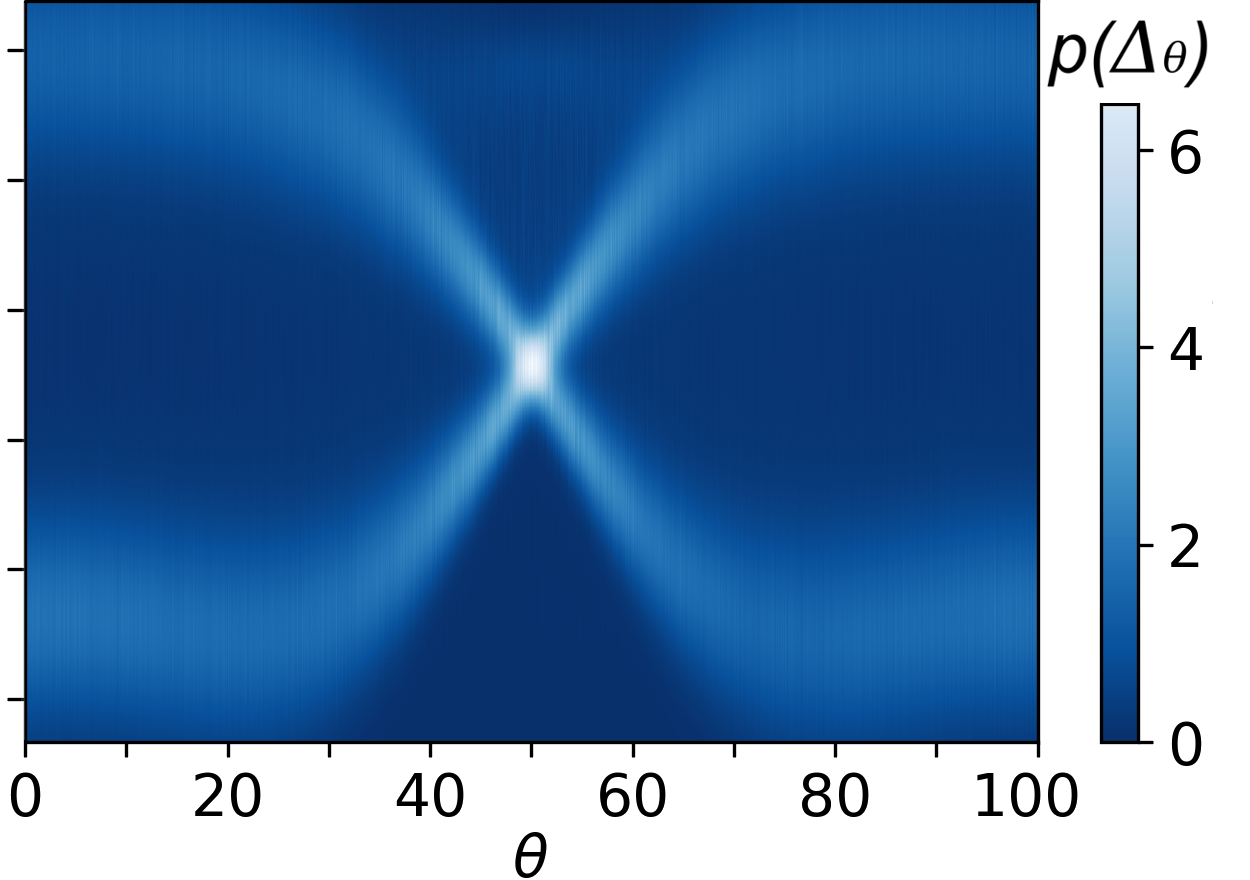}} 
    \caption{\textbf{(a)} An example of a multimodal target distribution: the discrepancy $\dt$ is bimodal for each value of the parameter $\bt$. \textbf{(b)} Vanilla GP as a surrogate distribution is unable to fit the target (red: observed data; line and shading: GP prediction with uncertainty). \textbf{(c)} Deep GP surrogate is able to accurately model the bimodal target distribution.} 
    \label{fig:multi-gp-dgp}
\end{figure*}

However, inferring simulator-based statistical models often requires approximating too complex distributions to be adequately represented by GPs, especially in the high-dimensional case. In particular, multimodal distributions \citep{shaw2007efficient, franck2017multimodal, li2021solving} still are a serious problem for the current LFI methods (Figure \ref{fig:multi-gp-dgp}).  Sequential neural density estimation methods, based on Masked Autoregressive Flows (MAFs) and Mixture Density Networks (MDNs) \citep{papamakarios2017, papamakarios2019}, use powerful deep network models to address this issue.  However, to our knowledge, no current method is flexible enough to handle multimodal target distributions, unless given numerous samples (beyond hundreds) which would, however, be infeasible for computationally costly simulators. Our research hypothesis is that by adopting highly flexible Deep Gaussian Processes (DGPs) as surrogates in BO, we can simultaneously model both uni- and multimodal target distributions, and further cover also non-stationarity and heteroscedasticity.

The main contributions are: 
\begin{itemize}
    \item We solve the LFI problem for multimodal target distributions, with a limited number of function evaluations, which is important for computationally heavy simulators.
    \item We propose quantile-based modifications for acquisition functions and likelihood approximation that are required for adopting Latent-Variable (LV) DGPs in BO for LFI. We provide a full computational complexity analysis for using LV-DGPs with these modifications.
    \item We give empirical evidence in several tasks, showing that the new surrogate model is able to handle well both unimodal and multimodal targets, as well as non-stationarity and heteroscedasticity. Consequently, the new DGP-based surrogate has a greater application range than vanilla GPs for solving LFI problems. We also show that the method outperforms alternatives that are based on neural density estimation and kernel mean embedding.
\end{itemize}

\section{Simulator-based inference}
\label{sec:2-sbi}

\begin{figure*}
    \centering
    \subfloat[Prior]{\label{fig:setting-1} \includegraphics[height=5cm]{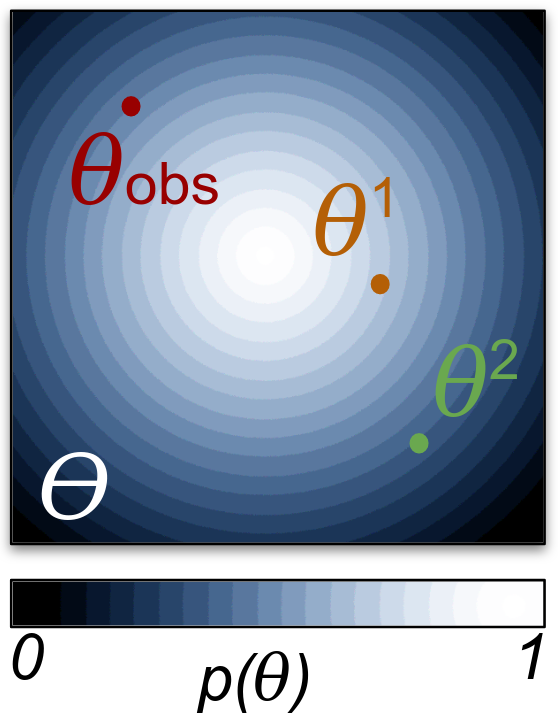}}
    \hfill
    \subfloat[Likelihood approximation through \eqref{eq:kdapprox} ]{\label{fig:setting-2} \includegraphics[height=5.cm]{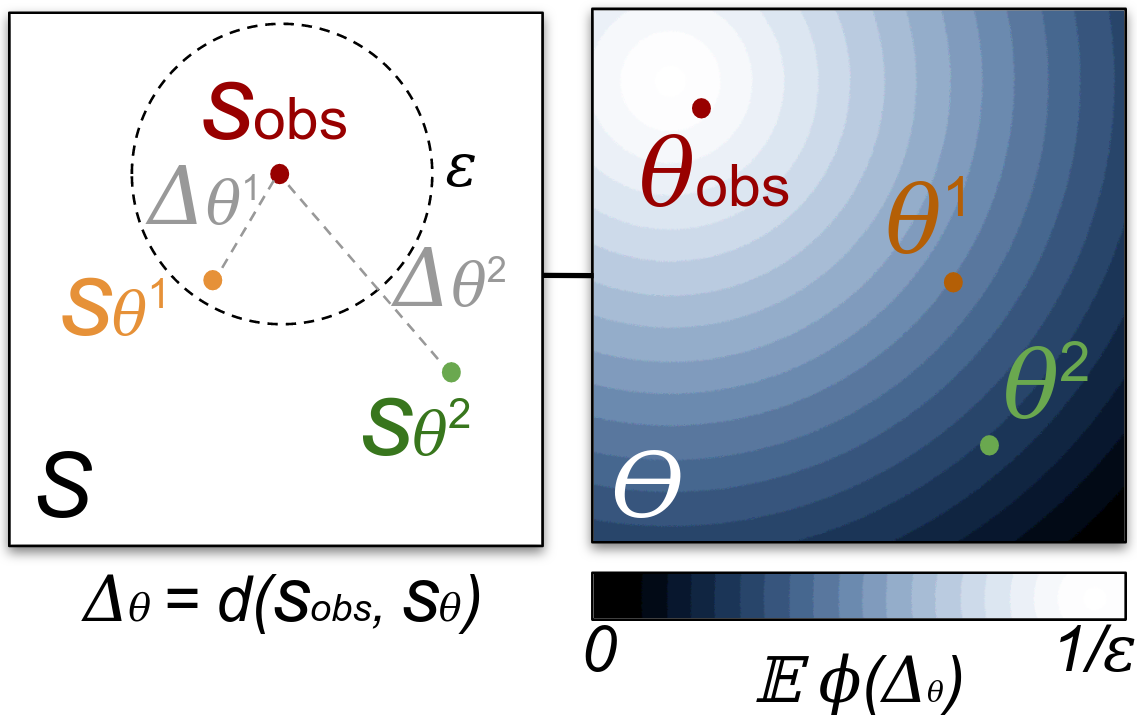}}
    \hfill
    \subfloat[Posterior]{\label{fig:setting-3} \includegraphics[height=5cm]{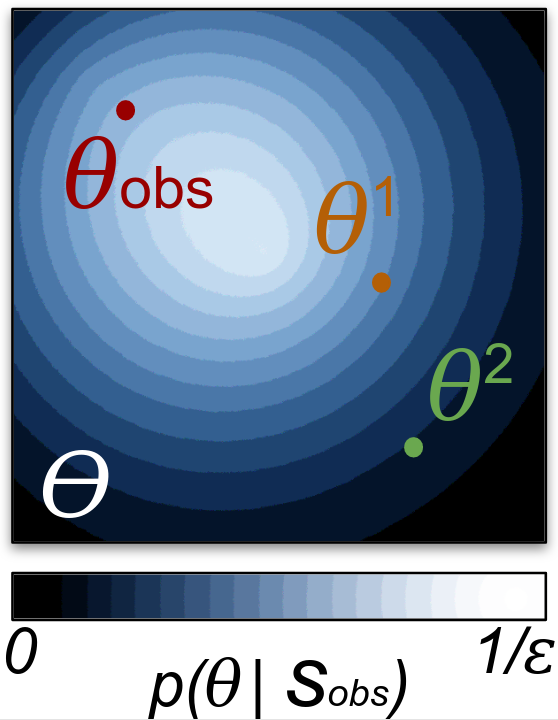}}
    \caption{Simulator-based inference estimates true parameters $\btobs$ of the observed dataset $\Xobs$. 
    (a) First, we assume the prior distribution over the parameter space. Parameter values, here $\bt^1$ and $\bt^2$, are selected to generate observations through a simulator that forms synthetic datasets $\Xthetafirst, \Xthetasecond$ (not in the figure).
    (b) Second, we replace the likelihood with an expectation over a kernel density \eqref{eq:kdapprox}. We use optional summarising functions to transform datasets back to a single point of summary statistics $\sthetafirst, \sthetasecond$. The discrepancies between datasets $\dtone$ and $\dttwo$ are measured (left) and used in a uniform kernel $\phi(\cdot)$ \eqref{eq:kernelfunc} to determine which parameters likely produced the observed dataset $\Xobs$. The expectation over uniform kernel density samples smooths the density surface, resulting in a likelihood approximation (right).
    (c) Third, we combine the prior and the likelihood approximation to compute the posterior $p(\bt | \sobs) \approx p(\bt | \Xobs)$. In the context of computationally expensive simulators, considered in this paper, each $\bt$ produces only one observation, and summaries serve as compact representations of observation points.}
    \label{fig:setting}
\end{figure*}

The general setting for simulator-based inference is illustrated in Figure \ref{fig:setting}. Our goal is to estimate $\btobs$ while only having the ability to draw simulated samples $\{\xtheta\} \sim p(\x | \bt)$. In Bayesian inference, we want to estimate the posterior distribution of $\bt$ instead of a point estimator. This work uses the surrogate model approach \citep{gutmann2016bayesian} to LFI with BO \citep{shahriari2016}.

\subsection{Approximate Bayesian Computation}
\label{sec:2-abc}

In LFI, the target likelihood $p(\Xobs|\bt)$ of the observed data $\Xobs$ given estimated parameters $\bt$ is implicitly modelled by a stochastic simulator, when its analytical form is unavailable. Arguably the most popular approach which has been almost synonymous to LFI is ABC. In ABC, the inference of the unknown parameter value that generated $\Xobs$, is based on quantifying the discrepancy $d$ between the summarised observed and synthetic datasets,
\begin{align}
\label{eq:discr}
    \dt &:= d\big[ \underbrace{\s(\Xobs)}_{\sobs}, \underbrace{\s(\Xtheta)}_{\stheta} \big] \ge 0.
\end{align}
Here, $d(\cdot,\cdot)$ is a metric scalar distance (e.g.~Euclidean distance) and $\s(\cdot)$ is a summarising function of the synthetic and observed datasets. Summary statistics are used to obtain a low-dimensional approximation of the likelihood,
\begin{align}
    p(\Xobs | \bt) &\approx p(\sobs | \bt),
\end{align}
which still inherits the intractability of the true likelihood. The discrepancies \eqref{eq:discr} are used with a kernel density estimate $\phi(\dt)$ (Figure \ref{fig:setting-2}) to approximate the summary likelihood \citep{sisson2018},
\begin{align}
\label{eq:kdapprox}
    p(\sobs| \bt) &\approx \E_{\Xtheta \sim p(\X | \bt)} \big[ \phi(\dt) \big].
\end{align}

A common choice for the kernel function $\phi(\cdot)$ is a uniform kernel \citep{sisson2018}:
\begin{align}
\label{eq:kernelfunc}
    \phi(\dt) = \left\{ \begin{array}{cc}
         \frac{1}{\epsilon}, & \quad \dt \in [0, \epsilon),   \\
         0, & \quad \text{otherwise},
    \end{array} \right.
\end{align}
where $\epsilon$ is the user-defined tolerance for the discrepancy. The kernel function $\phi(\cdot)$ quantifies the variability tolerance for simulated datasets. As a result, the approximate likelihood in \eqref{eq:kdapprox} becomes proportional to the empirical probability of the discrepancy being below the threshold $\epsilon$.

Finally, once the likelihood has been approximated, the Bayesian posterior over the parameter $\bt$ can be inferred through (Figure \ref{fig:setting-3})
\begin{align}
\label{eq:post}
    p(\bt | \sobs) &\propto p(\sobs | \bt) p(\bt).
\end{align}

In this work, we use the approximate likelihood $p(\sobs | \bt)$ in an importance-weighted sampling procedure to weight the posterior samples and calculate $p(\bt | \sobs)$. For a more detailed review of ABC methods and their recent advances, see \citet{lintusaari2017fundamentals, sisson2018}.

\subsection{Surrogate models in likelihood-free inference}
\label{sec:2-surrogates-lfi}

Among the first surrogate model based solutions for the LFI problem were the synthetic likelihood approaches, where the simulator output is approximated with a Gaussian distribution. \citet{wood2010} generated several $\xtheta$, or $\stheta$, at the parameter value $\bt$, and then used them to estimate the mean and covariance of the Gaussian. Synthetic likelihoods can be formulated in the Bayesian framework \citep{price2018}, which allows incorporating prior beliefs and updating them when new data are observed. GPs also lend themselves well to surrogate-modelling in LFI in multiple ways. \citet{meeds2014} used GPs as a surrogate for the proposal distribution in Markov Chain Monte Carlo ABC, and \citet{gutmann2016bayesian} modelled the discrepancy function as a function of the unknown parameters with a GP. 

Sequential neural density estimators and kernel mean embedding methods are recent surrogate model approaches to the LFI problem. Masked Autoregressive Flows (MAFs) \citep{papamakarios2017} and Mixture Density Networks (MDNs) \citep{papamakarios2019} use deep network models, resulting in accurate density estimations that have been suggested to require only ${O}(10^2)-{O}(10^3)$ samples for training. On the other hand, kernel mean embedding approaches tackle the problem of providing an embedding of the synthetic dataset to a reproducing kernel Hilbert space, removing the need of finding sufficient summary statistics \citep{nakagome2013kernel} or automatically tuning the threshold $\epsilon$ parameter \citep{hsu2019bayesian}. These approaches are yet to be used in BO to improve sample-efficiency further, as suggested by \cite{nakagome2013kernel, hsu2019bayesian}; we provide the first comparisons in Section \ref{sec:4-exp}. Our results show that both sequential neural density and kernel mean embedding approaches do not match the best methods with ${O}(10)-{O}(10^2)$  sample sizes, and more research into their representativity versus training cost is still needed.

\subsection{Bayesian optimisation for likelihood-free inference}
\label{sec:2-bolfi}

The task of approximating the likelihood $p(\sobs | \bt)$ can be formulated as an optimisation problem. Given the kernel function \eqref{eq:kernelfunc}, the expectation \eqref{eq:kdapprox} is maximised when simulated datasets $\Xtheta$ produce discrepancy below the threshold $\epsilon$. Thus, we need to optimize $\bt$ to minimize the distance $\dt$. To minimize the number of sampled datasets, we turn to BO, which has earlier been applied to LFI in a model called BOLFI \citep{gutmann2016bayesian}. 

BO requires a surrogate model for the objective and an acquisition function to guide the optimisation process. In BOLFI, the discrepancy \eqref{eq:discr} is the objective, and it is approximated with a Gaussian process (GP) \citep{williams2006gaussian} surrogate
\begin{align}
    \dt \sim GP(m(\bt), k(\bt,\bt')),
\end{align}
which defines the prior mean and covariance of the discrepancy surface:
\begin{align}
    \E[\dt]&= m(\bt) \\
    \cov[ \dt, \dt'] &= k(\bt,\bt').
\end{align}

The GP definition implies that any finite set of distances $\dt^{1:N} = \{\dt^n \}_{n=1}^N$ at alternative parameter values $\bt^{1:N} = \{\bt^n\}_{n=1}^N$ is jointly Gaussian: 
\begin{align}
p(\dt^{1:N}) = N(\dt^{1:N} | \m, \K),
\end{align}
where $\m = \{m(\bt^n)\}_{n=1}^N \in \R^N$ and the kernel $\K \in \R^{N \times N}$ contains values $\K_{ij} = k(\bt^i,\bt^j)$. The commonly chosen RBF kernel induces smooth distance surfaces that allow efficient exploration. 

\begin{figure}
    \centering
    \includegraphics[width=0.55\textwidth]{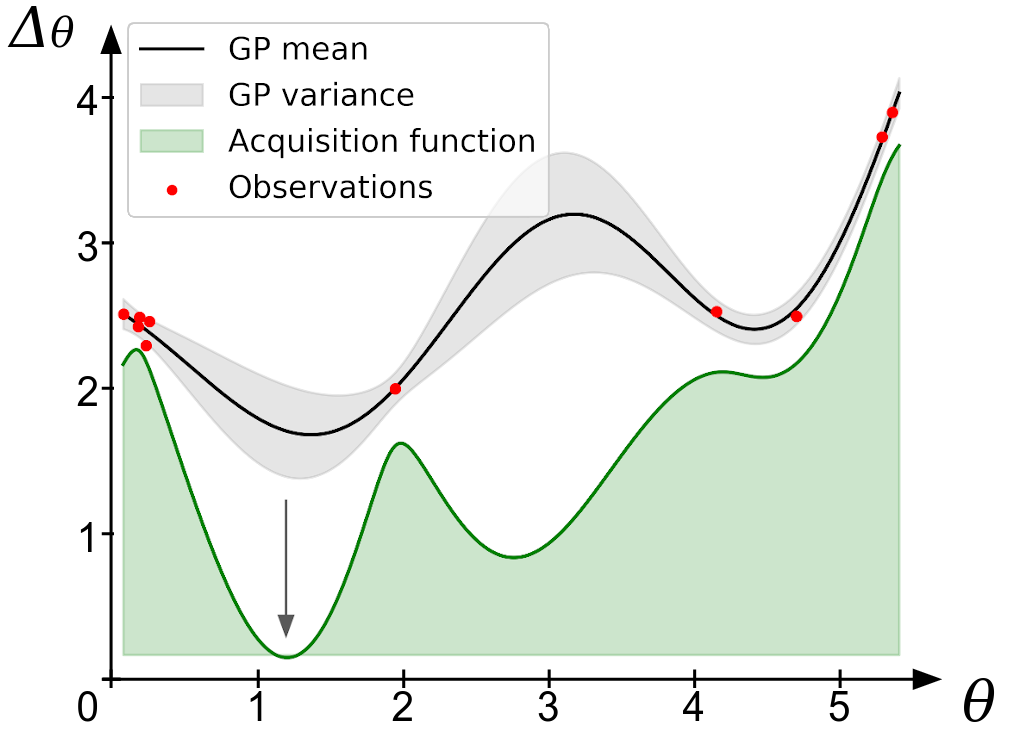}
    \caption{GP fit with the mean (black) and variance (grey) for 12 observations (red) collected through the BO procedure. The objective function will be sampled next at the minimum of the acquisition function (green), which is marked with an arrow.}
    \label{fig:bodgp}
\end{figure}

BO chooses the point $\bt^{t+1}$ where to next evaluate the objective function by minimising an acquisition function $\A^t(\cdot)$, such as lower confidence bound \citep{cox1992statistical}, at time $t$
\begin{align}\label{eq:acquisition}
\bt^{t+1} &= \text{arg min}_{\theta} \, \{ \A^t(\bt) \} \\
\A^t(\bt) &= m(\bt) - \sqrt{\eta_t^2 \cdot v(\bt)},
\end{align}
where $\eta_{t}^{2}$ is a user-defined tuning parameter \citep{srinivas2012information} and $v(\bt) := k(\bt, \bt)$ is the GP variance. The acquisition function uses the mean and variance of the GP, and is usually chosen to make a trade-off between exploitation (minimisation based on what is already known) and exploration (sampling in the regions of high uncertainty). See Figure \ref{fig:bodgp} for a demonstration.

Conventional BO works well for objectives with Gaussian uncertainties, but can be used with other surrogates as well. Some examples include: deep neural networks \citep{snoek2015scalable} for objectives that require many evaluations, DGPs \citep{hebbal2020bayesian} for non-stationary objectives, student-t processes \citep{shah2013bayesian} for modelling  heavy-tailed distributions, and decision trees \citep{jenatton2017bayesian} for modelling known dependency structures. 

In this work, we bring BO to solve a so-far unsolved problem: likelihood-free inference for commonly occurring irregular distributions, in particular multimodal but also skewed distributions. This is especially difficult for computationally heavy simulators, for which we can afford only few evaluations, and hence, need to adopt surrogate functions that combine flexibility with data-efficiency.

\section{Bayesian optimisation with deep Gaussian processes}
\label{sec:3-bodgp}

\begin{figure}
    \centering
    \includegraphics[width=0.7\textwidth]{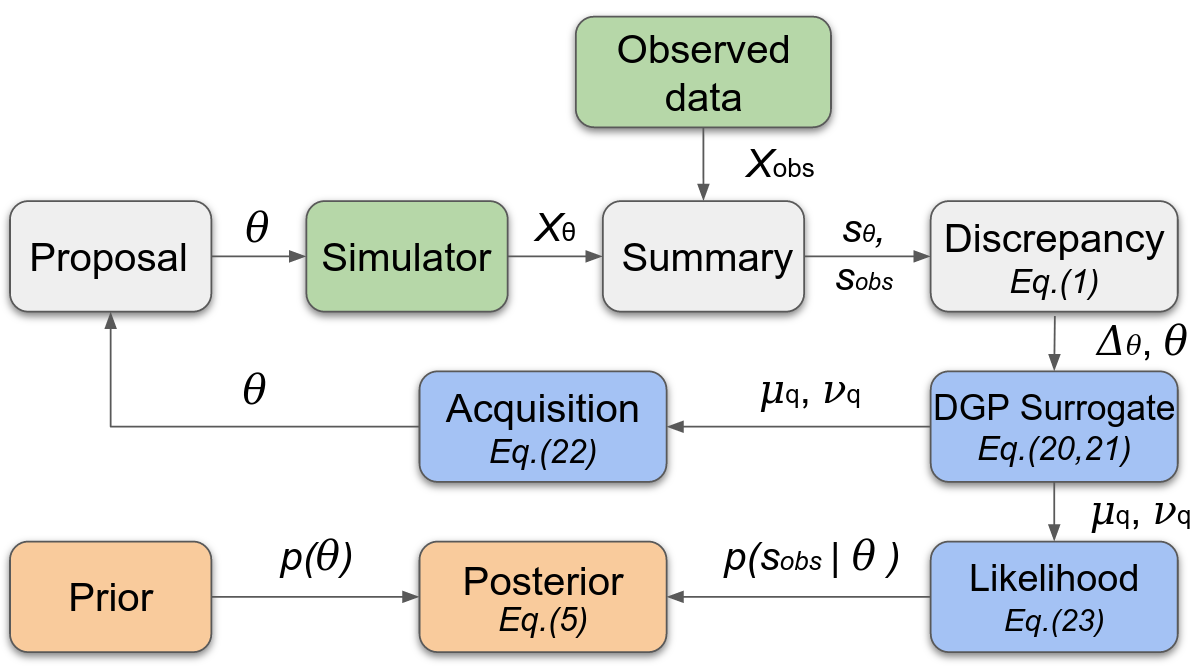}
    \caption{Overview of the proposed LFI with DGPs approach to estimating parameter posterior $p(\bt | \sobs)$. Given an observed dataset and a simulator (green blocks), we follow the BO procedure with the introduced DGP surrogate changes (blue blocks). Each parameter proposed by the acquisition function is run through the simulator to obtain a single synthetic dataset $\Xtheta$. The discrepancy $\dt$ is then computed using the summaries $\sobs$ and $\stheta$ for the observed and simulated datasets with \eqref{eq:discr}, and coupled with the corresponding $\bt$ to form the evidence for training the surrogate. Finally, the likelihood approximation $p(\sobs | \bt)$ is extracted and used along with the prior $p(\bt)$ to infer the posterior $p(\bt | \sobs)$.}
    \label{fig:approach}
\end{figure}

BO uses a probabilistic surrogate to find the posterior distribution of the parameter $\bt$. We propose to use DGP surrogates that are capable of handling multimodal and non-stationary discrepancy distributions (Section \ref{sec:3-dgps}), along with quantile-based modifications for an acquisition function (Section \ref{sec:3-bodgp-sub}) and likelihood approximation (Section \ref{sec:3-likelihoodapprox}) required for modelling such distributions in BO for LFI. In Section \ref{sec:3-complexity}, we evaluate the computational overhead from the new surrogate. The general overview of the proposed approach is illustrated in Figure \ref{fig:approach}.

\subsection{Multimodal deep Gaussian processes}
\label{sec:3-dgps}

A DGP composes multiple GPs together for more flexible and powerful function representations \citep{damianou2013deep,dunlop2018deep}. These representations can have a non-Gaussian, multimodal distributional form. However, DGP posteriors do not have explicit analytical solutions as GPs, and require variational \citep{salimbeni2017doubly} or Monte Carlo \citep{havasi2018inference} approximations for inference. 

The quality of the predictive posterior approximation largely depends on the chosen inference method, and most DGP models and inference methods are not able to yield multimodal marginals. Such irregularly behaved distributions can be modelled only when DGP latent function values $\f$ do not correlate with each other for the same input. In this work, we argue that it is important to use one of the possible solutions that guarantees this property. In the experiments, we have used Latent-Variable (LV) DGPs \citep{salimbeni2019deep}, which augment the input vector $\x$ with latent variables $w \sim N(0,1)$ concatenated into $[\x,w] \in \R^{D+1}$, to be used as input for the next GP layer. By combining different LV and GP layer architectures with importance-weighted objectives, LV-DGPs provide more flexible DGP posterior approximations \citep{salimbeni2019deep}. 

The proposed method in this paper is not exclusive to the specific LV-DGP model, and can be used with any DGPs that are capable of approximating multimodal marginal distributions. Therefore, we refer to LV-DGPs when we discuss this specific architecture, and to DGPs, whenever the results apply to multimodal capable DGPs in general. The specific LV-DGP method we used in the experiments is as follows; here with a LV layer followed by two GPs and denoted by LV-2GP. The GP prior is 
\begin{align}
\label{eq:dgpf}
    f(\x) &\sim GP(m_f(\x), k_f(\x, \x') ) \\
    g(\x) &\sim GP(m_g(\x), k_g(\x, \x') ),
\end{align}
and with Gaussian likelihoods for modelling the discrepancy \eqref{eq:discr}
\begin{align}
p(\dt | f, g, w) &= N(\dt | f(g([\bt, w])), \sigma^2) \label{eq:dgplik} \\
p(\dt | f, g) &= \E_{p(w)}  N(\dt | f(g([\bt, w])), \sigma^2) \label{eq:dgpexp}.
\end{align}

We use the Importance-Weighted Variational Inference (IWVI) of \citet{salimbeni2019deep} to minimize KL-divergence $\kl[q(\f, \g, \w) || p(\f, \g, \w | \dt)]$, where the $q(\f,\g,\w)$ are the variational posterior approximations to be learned. The lower bound can be formulated as
\begin{align}
    \text{log } p(\dt) &\geq \E_{q(\f,\g,\w)}  \text{log } p(\dt^{1:N} | \f, \g, \w) - \kl\big(q(\f,\g,\w) || p(\f,\g,\w)\big),
\end{align}
where we assume factorised variational approximation and prior,
\begin{align}
    q(\f,\g,\w) &=  q(\f) q(\g) q(\w) \\
    p(\f,\g,\w) &=  p(\f) p(\g) p(\w).
\end{align}
The latent variables further factorise as
\begin{align}
    q(\w) &= \prod_n N(w_n | a_n, b_n),
\end{align} 
with $a_n$, $b_n$ being variational parameters to be optimised. The variational approximations $q(\f),q(\g)$ represent Gaussian process layers, for which we use the sparse inducing point approximation \citep{salimbeni2019deep}. Later in the experiments, we use deeper architectures that use up to five GP layers. Similarly as in the two-layer example, these additional layers require inference of corresponding KL terms to form a composite function in \eqref{eq:dgplik} and \eqref{eq:dgpexp}. Once we have a DGP predictive distribution, we can use it in BO.

\subsection{Bayesian optimisation with deep Gaussian processes}
\label{sec:3-bodgp-sub}

BO requires a surrogate model, an acquisition function $A^t(\bt)$ and the ability to evaluate the black-box objective function. Here, we minimize the discrepancy $\dt$ as the objective and use the DGP probabilistic model in the acquisition function of the discrepancy $\dt$ from \eqref{eq:dgpexp} to choose where to sample next.

In BO for LFI, BO uses GP predictive mean and variance in the acquisition function, as shown in \eqref{eq:acquisition}. By design, the acquisition function focuses on accurate representation of the low-valued discrepancy regions. However, when the discrepancy is multimodal, the GP mean and variance tend to overestimate the expected discrepancy value and its uncertainty. As a result, multimodal and  more promising regions can be overlooked by BO in favour of unimodal regions. The solution to this problem is to accurately represent the low-valued regions of discrepancy, maintaining high signal-to-noise ratio. 

We introduce quantille-conditioning on DGP predictive samples to estimate more accurately the lowest values of the discrepancies. The DGP is applied to the regression problem $\bt \mapsto \dt$, resulting in estimates of mean $\mu_q(\bt)$ and variance $\nu_q(\bt)$ in the lowest quantile through a quantile function $Q(\cdot)$
    \begin{align}
    \mu_q(\bt) &= \E \{ \dt^n: \dt^n \leq Q(\epsilon_q)\}_{n=1}^N \label{eq:dgpqmean} \\
    \nu_q(\bt) &= \var \{\dt^n: \dt^n \leq Q(\epsilon_q)\}_{n=1}^N. \label{eq:dgpvar}
\end{align}

\begin{figure*}
    \centering
    \subfloat[All GP predictive samples]{\label{fig:qcon-1} \includegraphics[height=6.cm]{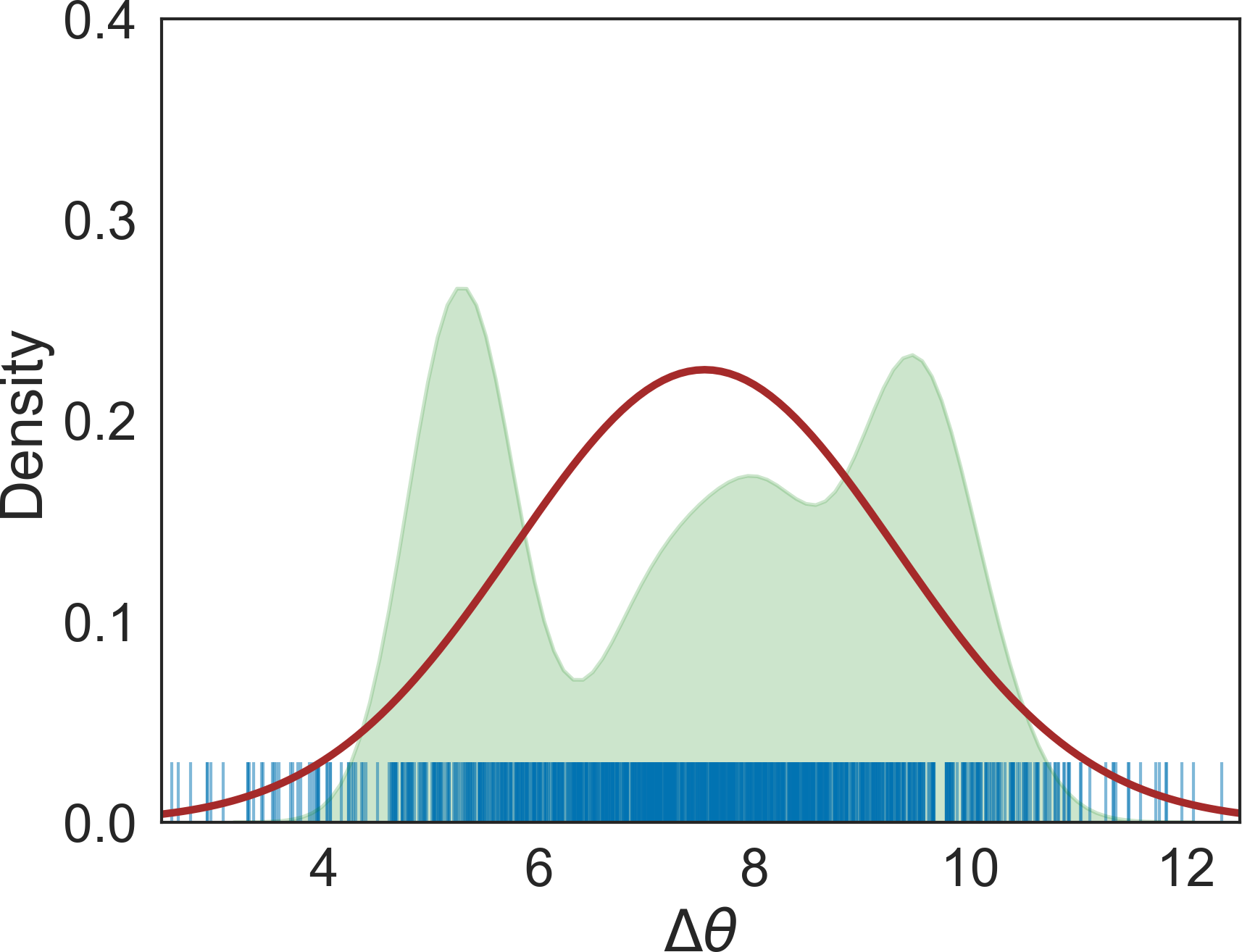}}
    \subfloat[GP predictive samples after quantile-conditioning]{\label{fig:qcon-2} \includegraphics[height=6.cm]{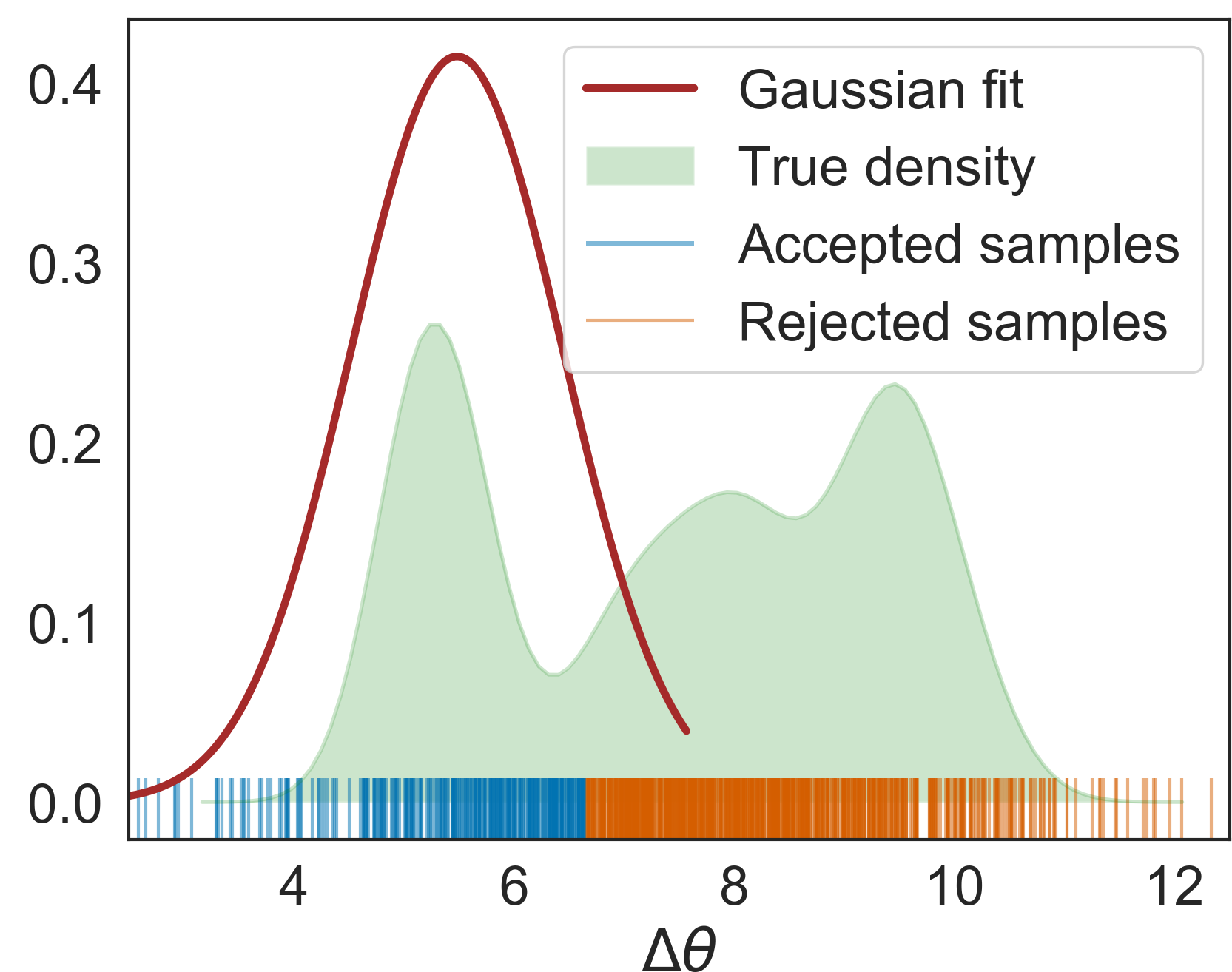}}
    \\
    \subfloat[All DGP predictive samples]{\label{fig:qcon-3} \includegraphics[height=6.cm]{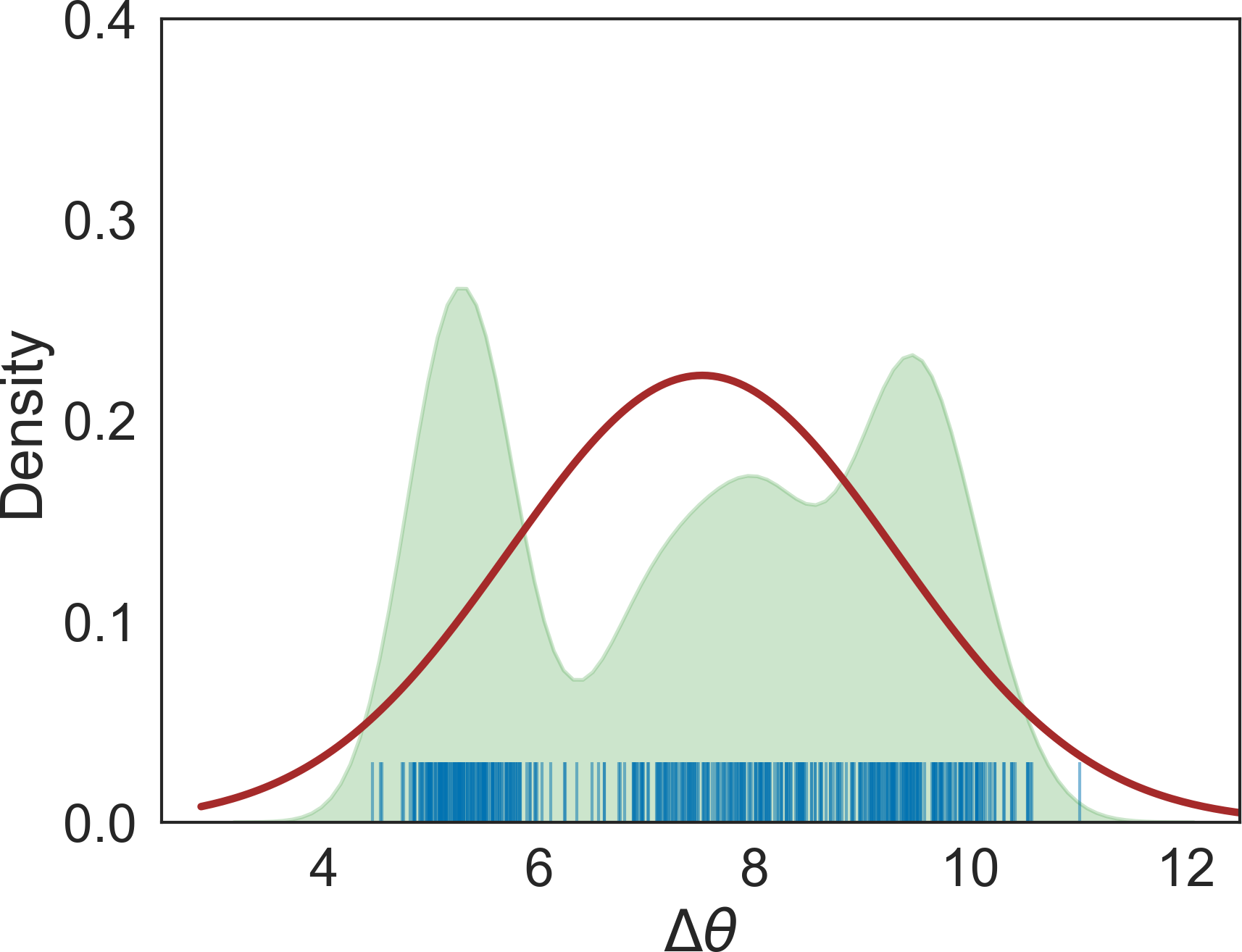}}
    \subfloat[DGP predictive samples after quantile-conditioning]{\label{fig:qcon-4} \includegraphics[height=6.cm]{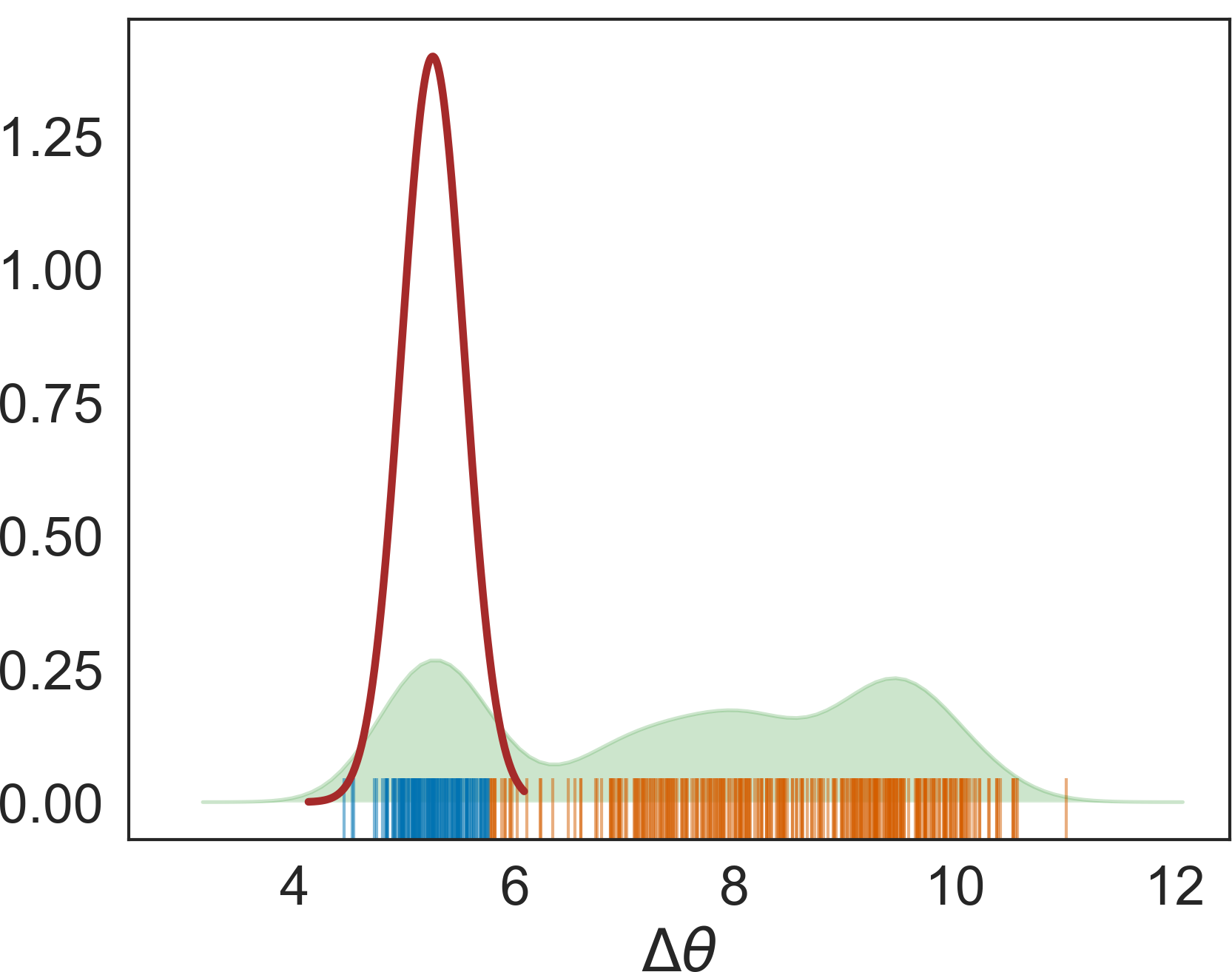}}
    
    \caption{Gaussian fit (red) before and after applying quantile-conditioning on GP and DGP predictive samples for the true density (green). GP predictive samples (blue) tend to overestimate the uncertainty of the true density (green) for the whole distribution (a) and for the lowest quantile (b) (with $\epsilon_q = 0.3$). At the same time, DGP predictive samples (blue) exhibit similar behaviour on the whole distribution (c), but with quantile-conditioning (d) result in a more accurate and narrow approximation of the low-valued discrepancy $\dt$ region, characterized as a Gaussian with the mean $\mu_q$ and variance $\nu_q$. This can be seen by comparing how closely the Gaussian curve (red), built on top of samples below the quantile-threshold (blue), estimates the low-valued discrepancy region of the true density (green). Predictive samples that are above the quantile-threshold are marked with orange. }
    \label{fig:qcon}
\end{figure*}

By only considering discrepancies below a (user-defined) small quantile-threshold $\epsilon_q$ (called \emph{quantile-conditioning} below), the estimator is able to focus on accurately representing the important low-valued regions of the discrepancy surface, as demonstrated in Figure \ref{fig:qcon}. We use these values in the acquisition function $\A^t(\bt)$ to get a new proposal for simulation, resulting in a simple quantile-based modification of the lower confidence bound selection criterion (LCBSC) \citep{cox1992statistical} for selecting a new parameter point $\bt^{t+1}$ at any current time $t$
\begin{align}\label{eq:qacquisition}
\A^t(\bt) &= \mu_q(\bt) - \sqrt{\eta_t^2 \cdot \nu_q(\bt)},
\end{align}
The proposed quantile-based acquisition maintains the advantages of the LCBSC, while also enabling BO with multimodal or skewed uncertainties.

\subsection{Likelihood approximation}
\label{sec:3-likelihoodapprox}

Lastly, we use the mean and the variance of DGP posterior samples below a quantile $\epsilon_q$ threshold to approximate the likelihood \eqref{eq:kdapprox}. \citet{gutmann2016bayesian} constructed the likelihood approximation from the GP model of the discrepancy using normal cumulative distribution function (cdf) with the discrepancy tolerance $\epsilon$. This approach works well for unimodal distributions, where the mean and variance of the GP characterize the mode well, but for multimodal distributions individual modes are concealed when represented with the mean and variance of the whole distribution. Moreover, only the modes that correspond to low-valued discrepancy regions are likely to produce the observed dataset and, hence, should be considered in the likelihood approximation. Here, we filter out all DGP predictive posterior samples that are above quantile-threshold to focus on samples from the low-valued discrepancy regions:

\begin{align}
    \label{eq:likapprox}
    p(\sobs | \bt) &\propto F\left(\frac{\epsilon - \mu_q(\bt)}{\sqrt{\nu_q(\bt) + \sigma^2}} \right),
\end{align}
where $F(\cdot)$ is the cdf of Gaussian with mean 0 and variance 1, $\mu_q$ and $\nu_q$ are the mean and the variance of DGP posterior sample below the quantile-threshold $\epsilon_q$, $\epsilon$ is a tolerance from \eqref{eq:kernelfunc} and $\sigma^2$ is the Gaussian likelihood noise from \eqref{eq:dgpexp}. As the number of predictive samples grows and the quantile-threshold $\epsilon_q$ is lowered, this approximation becomes more accurate. 

In summary, we have introduced a way for DGP surrogates to handle irregularly-behaved marginal distributions in the context of BO, by proposing a quantile-based likelihood approximation and acquisition rule.

\subsection{Computational overhead}
\label{sec:3-complexity}

The computational overhead of having a more complex surrogate is negligible if the simulator is computationally expensive. DGPs, as a more flexible model, require more time for training and prediction, compared to GPs. There are three major stages of the BOLFI algorithm \ref{alg:bolfi}, where the surrogate plays a role: initialisation, BO updates and posterior extraction. In this section, we analyse the increase in time complexity caused by switching to multimodal capable LV-DGPs instead of traditionally-used GPs at every stage. We consider a LV-DGP architecture with $l$ GP layers, introduced in Section \ref{sec:3-dgps}, and sparse approximations of GPs in our analysis.

\begin{algorithm}
\SetAlgoLined
\KwData{Datasets $\xobs$, $N$ initial simulation budget, $S$ BO simulation budget}
\KwResult{Posterior $p(\bt | \xobs)$}
    sample $N$ times from the prior $\bt^{1:N} \sim p(\bt)$\;
    simulate synthetic datasets $\Xtheta^{1:N} = g(\bt^{1:N})$\;
    compute discrepancies $\dt^{1:N}$, Eq \eqref{eq:discr}\;
    
    initialize DGP as $\bt^{1:N} \mapsto \dt^{1:N}$\;
    train DGP with pairs $\{\bt^{1:N}, \dt^{1:N}\}$\;
 
    starting BO procedure\;
    \While{current simulation budget $< S$}{
    acquire new $\bt'$, Eq \eqref{eq:qacquisition}\;
    simulate new datasets $\xtheta' = g(\bt')$\;
    compute discrepancy $\dt' $, Eq \eqref{eq:discr} \;
    augment DGP data with $\{\bt', \dt'\}$\;
    }
    
    retrain DGP\;
    extract result (find DGP minimum value)\;
    extract posterior $p(\bt | \xobs)$, Eq \eqref{eq:likapprox}\;
 \label{alg:bolfi}
 \caption{BOLFI algorithm with DGPs}
\end{algorithm}

\paragraph{Initialisation.} At this stage of the algorithm, the simulator creates initial observations and trains the surrogate model. Sparse GPs require ${O}(m^2 n  d_\theta)$ cost for inference, and ${O}(m d_\theta)$ and ${O}(m^2 d_\theta)$ for predicting the mean and variance respectively. Here, $m$ is the number of inducing points, $n$ is the number of initial data, and $d_\theta$ is the dimensionality of a parameter vector. LV-DGPs, on the other hand, are using sample average of $k$ terms (importance-weight samples) to replace the latent variable layer, resulting in ${O}(l k m^2 n d_\theta)$ for training, and ${O}(l k m d_\theta)$ and ${O}(l k m^2 d_\theta)$ for the mean and variance prediction.

\paragraph{BO.} Once the surrogate model has been trained, the BO procedure starts. It consists of minimisation of the acquisition function, simulation of data, and optional surrogate model hyper-parameter optimisation or retraining (in our implementation, we do this last step at the final stage). Again, when the simulator is fast, the acquisition function minimisation becomes the computational bottleneck of this stage. We used L-BFGS-B optimisation \citep{zhu1997algorithm} for finding a minimum of the acquisition function \eqref{eq:qacquisition} with the cost ${O}(t  d_\theta  \A  i)$, where $t$ is the number of steps stored in memory by parameter declaration (the limited memory BFGS method does not store the full hessian but uses this many terms in an approximation to it), $i$ is the number of initialisation points and $\A$ is the cost of the acquisition function call. Further decomposition of the acquisition function complexity requires computation of DGP mean and variance, bounded by ${O}(l k m^2 d_\theta)$ cost, corresponding to the DGP predictive variance, or ${O}(p^2)$, corresponding to calculation of the quantile-conditioning based on $p$ predictive samples. Finally, when the acquisition function minimum is found, the points in the batch are calculated by adding an acquisition noise to its value.

\paragraph{Posterior extraction.} As the final stage, the posterior is extracted from the trained surrogate. This is performed by sampling the prior $S$ times, calculating the threshold $\epsilon$, and then reweighing the samples by using \eqref{eq:likapprox}. The prior sampling and importance weighted resampling are less computationally intensive than finding the threshold and calculating the weights, where DGP prediction plays an important role. The threshold is found by minimising the DGP mean function with L-BFGS-B minimisation that requires ${O}(t  d_\theta  \mu  i)$ cost, where $\mu$ is the cost of the DGP mean function ${O}(l k m d_\theta)$. And applying Equation \eqref{eq:likapprox} has complexity of $O(l k m^2 d_\theta S)$ or ${O}(p^2)$, since it requires calculating DGP predictive mean and variance for $S$ samples conditioned on the quantile-threshold $\epsilon_q$.

In summary, the increase in complexity from switching GPs to LV-2GPs is ${O}(l k)$ times for all three stages. In practice, $k$ is a relatively small number ranging from 5 to 20 and $l$ is from 2 to 5. Additionally, the cost of calculating the quantile-threshold should rarely exceed the cost of calculating predictive mean and variance with predictive samples, ranging from 10 to 100.

\section{Experiments}
\label{sec:4-exp}

We study the merits of DGP surrogates in BO first in illustrative demonstrations and then in two case studies. Our main goal is to reduce the number of required simulator evaluations, which is important for computationally intensive simulators. Section \ref{sec:4-techdetails} describes the experimental setup, simulators and comparison methods that were used. In Section \ref{sec:4-gpvsdgp}, we consider the simplest deep LV-2GP architecture of the LV-DGP model and analyse its advantages against traditionally used GPs. In Section \ref{sec:4-addexp} we compare LV-DGP architectures with neural density estimation and kernel mean embedding approaches. The results of the experiments are summarised in Table \ref{tab:wd} with details of the findings discussed in individual sections afterwards.

\begin{table*}
    \centering
    \caption{LV-DGP models showed the best results in four out of five test cases (columns) across all alternative models (rows). In all case studies, both DGPs and GPs produced better approximations of the posterior than the other models. Only in the BDM case, MDN outperformed DGP (but not GP), and in TE2 MAF outperformed GPs (but not DGPs). The performance was measured with 95\% confidence interval (CI) of the scaled Wasserstein distance between the surrogate model posterior and the true posterior of $\theta$, across 1000 runs. Models from Section \ref{sec:4-gpvsdgp} and the best results in each column are highlighted in bold. * denotes models that used 1000 total observations instead of 200 for the sample-efficiency comparison.}
    \centerline{
    \begin{tabular}{c c c c c c}
        \\
        \hline
        {\bf Model} & {\bf TE1} & {\bf TE2} & {\bf TE3} & {\bf BDM} & {\bf NW} \\ \hline
        2GP & (1.84, 1.9) & (2.1, 2.22) & (1.19, 1.2) & (1.51, 1.59) & (1.33, 1.35) \\ 
        4GP & (1.86, 1.92) & (2.03, 2.06) & {\bf (1.18, 1.19)} & (1.49, 1.57) & (1.31, 1.33) \\ 
        
        LV-GP & (1.93, 2.0) & {\bf (1.53, 1.57)} & (1.37, 1.41) & (1.39, 1.46) & (1.4, 1.45) \\
        
        \textbf{LV-2GP} & (1.83, 1.89) & (1.6, 1.64) & (1.23, 1.26) & (1.51, 1.61) & {\bf (1.24, 1.29)} \\ 
        LV-3GP & {\bf (1.82, 1.88)} & (1.68, 1.72) & (1.23, 1.25) & (1.47, 1.55) & (1.25, 1.29) \\ 
        LV-4GP & (1.85, 1.9) & (1.7, 1.74) & (1.22, 1.24) & (1.5, 1.6) & (1.26, 1.29) \\ 
        LV-5GP & (1.86, 1.92) & (1.74, 1.77) & (1.21, 1.23) & (1.47, 1.54) & (1.27, 1.3) \\ 
        LV-6GP & (1.86, 1.91) & (1.75, 1.79) & (1.2, 1.22) & (1.47, 1.54) & (1.27, 1.3) \\ 
        \textbf{GP} & (1.89, 1.95) & (2.65, 2.68) & (1.2, 1.21) & {\bf (1.23, 1.25)} & (1.67, 1.7) \\
        MAF & (10.44, 14.47) & (1.99, 2.02) & (62.71, 84.58) & (2.03, 2.16) & (2.37, 2.5) \\ 
        MAF* & (13.66, 18.45) & (2.02, 2.04) & (59.13, 79.92) & (1.79, 1.88) & (2.29, 2.42) \\ 
        MDNs & (8.66, 11.7) & (15.63, 18.16) & (14.5, 22.28) & (1.38, 1.4) & (1.8, 1.83) \\ 
        MDNs* & (12.95, 17.62) & (29.37, 34.6) & (36.35, 51.16) & (1.38, 1.4) & (1.8, 1.84) \\ 
        KELFI* & (24.18, 32.01) & (2.06, 8.7) & (2.29, 3.31) & (18.48, 19.74) & (6.66, 6.98) \\ \hline
    \end{tabular}
    }
    \label{tab:wd}
\end{table*}

\subsection{Experimental setup}
\label{sec:4-techdetails}

In each simulation experiment, we select true parameter values, and use them to produce the observed data set with the simulator. Each experiment is repeated 1,000 times, the runs differing in the choice of random seeds that affect the observations used as initial evidence. We limit the number of total simulator calls to 200 with 100 initial evidence points drawn from the prior before the active learning procedure starts; when targeting computationally heavy simulators this is already plenty. In Section \ref{sec:4-gpvsdgp} we also study how the performance of the GP and DGP surrogates changes with fewer observations and initial evidence, where a half of all observations come from initial evidence points.

When evaluating goodness of the posterior approximations of $\bt$, we estimate the ground truth posterior numerically by Rejection ABC with $10^8$ simulations, and then select $0.1\%$ samples with the lowest discrepancy to represent the posterior distribution. Closeness of the estimated posterior $p_\text{sur}(\bt | \sobs)$ to this ground truth reference posterior $p_\text{ref}(\bt | \sobs)$ is measured with the empirical Wasserstein distance \citep{genevay2016stochastic} that shows similarity of the estimated surrogate posterior to the ground truth, defined as

\begin{align}\label{eq:wasserstein}
W_{\varepsilon}(\mu, \nu)=\int_{\X} u(\x) \mathrm{d} \mu(\x)+\int_{\Y} v(\y) \mathrm{d} \nu(\y)-\iota_{U_{c}}^{\varepsilon}(u, v)
\end{align}
where $\mu \in M_{+}^{1}(\X)$ and $\nu \in M_{+}^{1}(\Y)$ are two measures, defined on metric spaces $\X$ and $\Y$, $(u,v) \in C(\X) \times C(\Y)$ of ``ground cost'' space and $\iota_{U_{c}}^{\varepsilon}(u, v)$ is an indicator function. 
In our case, $\mu$ is the posterior of the surrogate model $p_\text{sur}(\bt | \sobs)$, and $\nu$ the ground truth posterior $p_\text{ref}(\bt | \sobs)$. We report the scaled (divided by the smallest value) Wasserstein distance $\W$. See \citet{genevay2016stochastic} for further details.

\subsubsection{Simulator descriptions}
\label{sec:4-simulators}

In our experiments, we used five different simulators, three of which are toy models and two are case studies. The toy models were designed to demonstrate specific properties of the surrogate model: non-stationarity, multimodality and heteroscedasticity. The case studies represent more difficult problems that often occur in practice. They have multi-dimensional parameters and cover both unimodal and multimodal cases. A more detailed description of each simulator is provided below.

\paragraph{Demonstrations: non-stationarity, multimodality and heteroscedasticity.} The discrepancy function of the first demonstrator TE1 is non-stationary with the ground truth $\thetaobs = 50$. The simulator function $g_\text{TE1}(\theta)$ generates data from the sum of three Gaussian density functions with different means and variances,

\begin{align}\label{eq:gte1}
g_\text{TE1}(\theta) = N(\theta | 30, 15) + N(\theta | 60, 5) + N(\theta | 100, 4) + \epsilon,
\end{align}
where $\epsilon \sim N(0,0.005)$.

The second toy example, TE2, has a multimodal discrepancy function with the ground truth $\thetaobs = 20$. The simulator function $g_\text{TE2}$ randomly `chooses' one of two logistic functions, and generates the observation according to
\begin{equation}
g_\text{TE2}(\theta^\prime) = \epsilon_2 \cdot \frac{\theta^\prime}{1 + \theta^\prime} + (1 - \epsilon_2) \cdot \frac{1}{1 + \theta^\prime} + \epsilon_1,
\end{equation}
where $\theta^\prime = \exp(-0.1 (\theta - 50))$, $\epsilon_1 \sim N(0, 0.01)$ and $\epsilon_2 \sim \mathsf{Bernoulli}(0.5)$. The simulator function creates several modes in the observation space, that later transfer to the discrepancy function.

Finally, the discrepancy function of the third demonstrator TE3 is heteroscedastic. The output of the simulator is generated as a sum of samples from two different beta distributions that are defined through the input parameter $\theta$. Similarly as in TE1, probability densities are used as basis functions:

\begin{align}\label{eq:posteriorapprox}
g_\text{TE3}(\theta) = \mathsf{Beta}(\theta + 1, 5) + \mathsf{Beta}(5, \theta + 1).
\end{align}

The ground truth of this case is $\thetaobs = 20$. A uniform prior on the interval $(0, 100)$ is used for simulator parameters, as in the first two demonstrators. For all three toy examples, the Euclidean distance is calculated directly on observations, since their dimension is low and there is no need for summary statistics.

\paragraph{Birth-Death model.} The Birth-Death model (BDM) describes tuberculosis transmission in the San Francisco Bay Area, as formulated by \citet{tanaka2006using}. Given epidemiological parameters $R_1, R_2, \beta, t_1$, the model simulates tuberculosis outbreak dynamics in a population and outputs cluster indexes of observed transmission cases. Our goal in inferring the BDM parameters is to approximate the posterior distribution $P(R_1, R_2, \beta, t_1 | \xobs)$, where $\xobs$ was generated with the vector of ground-truth parameters (5.88, 0.09, 192, 6.74). These parameter values were inferred by \citet{lintusaari2019resolving} from the summaries of real data \citep{small1994epidemiology}. We used the weighted Euclidean distance as the discrepancy measure with the summaries and the corresponding distance weights shown in Table \ref{tab:bdmss}. The same hierarchical priors as in \citet{lintusaari2019resolving} were used:

\begin{align}\label{eq:bdmpriors}
\theta_\text{burden} &\sim N(200,30) \\
\theta_{R_1} &\sim \mathsf{Unif}(1.01,20) \\
\theta_{R_2} | \theta_{R_1} &\sim \mathsf{Unif}(1.01, (1 - 0.05 \cdot \theta_{R_1})/0.95) \\
\theta_{t_1} &\sim \mathsf{Unif}(0.01, 30).
\end{align}

For detailed interpretation of simulator parameters and summaries, see \citet{lintusaari2019resolving}. 

\begin{table}
    \centering
    \caption{The summary statistics for the BDM case, their weights in the discrepancy function \citep{lintusaari2019resolving}}
    \begin{tabular}{p{0.35\textwidth} c}
        \hline
         \textbf{Summary} & \textbf{Weight} \\ \hline
         Number of observations & 1 \\ 
         Number of clusters & 1 \\ 
         Relative number of singleton clusters & 100/0.60\\ 
         Relative number of clusters of size 2 & 100/0.4\\
         Size of the largest cluster & 2 \\
         Mean of the successive differences in size        among the four largest clusters & 10 \\ 
         Number of months from the first observation      to the last in the largest cluster & 10 \\ 
         The number of months in which at least one  observation was made from the largest cluster & 10 \\ \hline  
    \end{tabular}
    \label{tab:bdmss}
\end{table}

\begin{figure*}
    \centering
    \subfloat[NW demonstration]{\label{fig:5-nwd}\includegraphics[height=5.5cm]{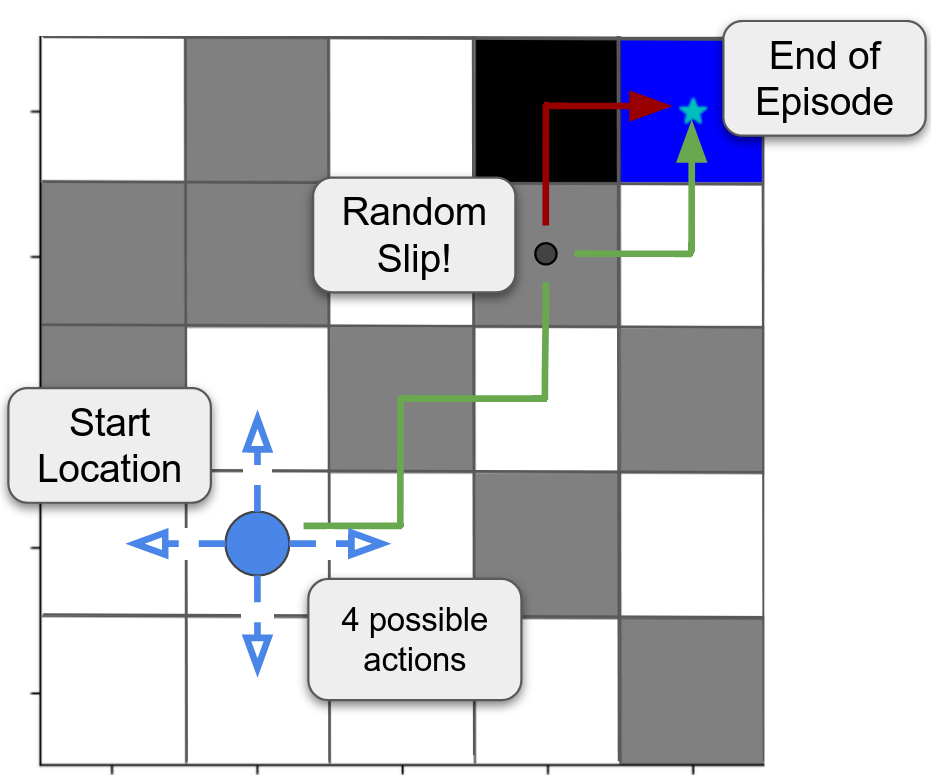}}
    \subfloat[NW experiment environment]{\label{fig:5-nwe}\includegraphics[height=5.5cm]{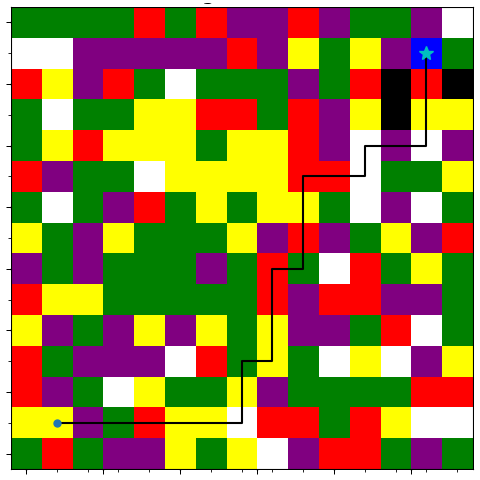}}
    \caption{(a) In the \textbf{NW} environment the agent (blue circle) starts at a fixed location and can perform four actions: going up, down, left and right. Since the environment is stochastic, the agent may deviate from the optimal green trajectory and end up in a black cell that heavily penalizes the reward. The episode ends once the agent reaches the goal. (b) The \textbf{NW} map that we used in the experiments, with an example observed trajectory shown. Our model needs to infer the reward cell colour parameters, given the summary statistics of the trajectory: in this case 9 turns, 24 steps, and 51 reward.}
    \label{fig:6-nw-map}
\end{figure*}

\paragraph{Navigation World.} The Navigation World (NW) model \citep{abel2019simple} is a simplified planning environment based on a grid world, where an agent needs to reach a target on a map (Figure \ref{fig:5-nwd}). The agent moves in four directions (up, down, left and right) and receives a reward based on the colour of the tile it visits (e.g. +100 for reaching the goal, -500 for the black cell). We formulate our inference task as an inverse-reinforcement learning problem, where the goal is to approximate the multidimensional distribution over the parameters of the Q-learning agent's \citep{littman1994markov} reward function operating on the NW map.

The NW agent learns in a stable environment, and then has to operate in a stochastic ``real world''. The agent always starts at a fixed position and explores the environment with no step cost. It is first trained on the map, and after a certain number of training episodes, we ask it to sample a trajectory (e.g. green trajectory in Figure \ref{fig:5-nwd}). However, this time, when we sample the trajectory, the agent can slip into an adjacent cell by accident (red trajectory in Figure \ref{fig:5-nwd}). This may lead to multiple distinct trajectories with different rewards, causing multimodality in the reward space. Moreover, the agent's policy may converge to multiple optimal solutions, depending on the training initialization, which also contributes to multimodality that causes multiple trajectories for the same parameter setting. In reinforcement learning, when an optimal policy or rewards have multiple modes \citep{kormushev2012simultaneous, barrett2008learning}, the solution to the inverse problem becomes particularly challenging \citep{franck2017multimodal, li2021solving}. 

The experiments were conducted on a more complex map, with tiles of five different colours corresponding to different rewards, shown in Figure \ref{fig:5-nwe}. The simulator starts by setting the reward parameters for each colour (five-dimensional vector), and then training the Q-agent for 8,000 episodes in a completely deterministic environment. Once the agent is trained, we sample 5 trajectories and learn their individual summaries: number of turns, number of steps and the reward.  For example, a trajectory with summaries (9, 24, 51) is illustrated in Figure \ref{fig:5-nwe}. We used the Euclidean distance between the summaries of the sampled and observed trajectories to fit the surrogate model, as well as independent uniform priors on the interval (-20, 0) for the simulator parameters, whereas true parameter values were (0.0, -1.0, -1.0, -5.0, -10.0).

\subsubsection{Comparison methods}
\label{sec:4-compmethods}

In this section, we specify the implementation details for the proposed LV-DGP surrogate in BOLFI and three comparison methods used in our experiments: GP-BOLFI, neural-density and kernel mean embedding. First, we describe LV-DGP surrogates, followed by GP surrogates that were traditionally used in BOLFI. Then, we introduce two variants of neural density estimators (MAF, MDN) that have performed the best in empirical LFI experiments earlier \citep{papamakarios2017, papamakarios2019}. We conclude with the implementation details for the Kernel Embedding for LFI (KELFI) method that outperformed other kernel mean embedding alternatives in \citet{hsu2019bayesian}. For all three comparison methods, we used the same hyper-parameters that were proposed in their original papers.

\paragraph{LV-DGPs.} We compared multiple architectures of the LV-DGP model, described in Section \ref{sec:3-dgps}. We use a naming convention where the name of the architecture specifies the exact sequence of layers, e.g. `LV-3GP' refers to a DGP with a LV layer followed by three GP layers. In all works, we used the squared exponential kernel. The initial value for the lengthscale before optimisation was set to the square root of the dimension, and the variance was fixed to 1, since the data was standardised. Kernel parameters and the likelihood variance (initialised with 0.01) were optimised from their initial values: the final layer using natural gradients (initial step size of 0.01) and the inner layers with the Adam optimizer (initial step size of 0.005) \citep{kingma2014adam}. Scaled conjugate gradient optimisation with the maximum number of function evaluations of 50 was used for GPs. In all experiments with LV-DGP models, we used $50$ inducing points, $5$ importance-weighted samples and $20$ samples for predictions and gradients. The quantile-thresholds $\epsilon_q$ for the acquisition function and the surrogate likelihood were set to $0.3$, so we get 5-30 posterior predictive samples after applying quantile-conditioning. These settings for the IWVI inference technique performed well across all experiments presented in the paper, and we recommend them as defaults parameters for the method. Comparison with Stochastic Gradient Hamiltonian Monte Carlo \citep{havasi2018inference} as an alternative inference method with convergence diagnostics can be found in the Supplementary material. We report results for $200$ total observations. The model was implemented in Python with GPFlow \citep{GPflow2017}. Engine for Likelihood-Free Inference (ELFI) \citep{elfi2018} was used as the platform for the implementations, and the proposed model is available in ELFI for application and further development (elfi.ai). The LV-GP \citep{wang2012gaussian, dutordoir2018gaussian} model is also used in the experiments. It shares the same input augmentation mechanism as the LV-DGP and implementation, but consists of only one GP layer.

\paragraph{GPs.} The vanilla GP surrogate was initially introduced for BOLFI \citep{gutmann2016bayesian}. The GP model had as hyperparameters the lengthscale of the squared-exponential kernel lengthscale, variance and added bias component. Gamma priors were used for all three of them, initialised by the expected value and variance chosen based on initial standardised data. We used LCBSC acquisition in BO. The model was implemented in Python with the GPy package \citep{gpy2014}.
 
\paragraph{MAF} \citep{papamakarios2017} is an implementation of normalising flow that uses Masked Autoencoder for Distribution Estimation (MADE) \citep{germain2015made} as building blocks, where each conditional probability is modelled by a single Gaussian component. In the experiments, we used the architecture with 5 stacked MADEs in the flow and 2 hidden layers, containing 50 hidden units (sequential strategy for assigning degrees to hidden nodes was used) with hyperbolic tangent as an activation function. The model was trained with Adam  optimisation, using a minibatch size of 100, and a learning rate of $10^{-4}$. L2 regularisation with coefficient $10^{-6}$ was added. The training was performed with 300 epochs in 5 batches, with the number of populations equal to the total number of observations divided by the number of batches. We report results for 200 and 1000 total observations.

\paragraph{MDN} \citep{papamakarios2019} is a feedforward neural network that takes the observation $\stheta$ as an input and outputs the parameters of a Gaussian mixture over $\bt$. We use an ensemble of 5 MDNs in our experiments with the same architecture: 2 hidden layers with 30 hidden units in each with the hyperbolic tangent activation function. The parameters for optimisation and training procedures were the same as for the MAF. We report results for 200 and 1000 total observations.

\paragraph{KELFI} \citep{hsu2019bayesian} is a surrogate  likelihood  model that leverages smoothness properties of conditional mean embeddings. Different strategies for adjusting marginal kernel means likelihood (MKML) hyperparameters were used for the experiments. For toy examples, hyperparameters $(\epsilon, \beta, \gamma)$ were chosen by gradient optimisation, denoted as `All-Opt' in the original paper. The initial values of $(0.06, 0.6, 10^{-6})$ were used for initialisation. However, for the BDM and the NW cases, we were unable to train the hyperparameters using gradients due to a numerical error in the original KELFI software. Therefore, for these cases we chose the grid based optimisation strategy, denoted as `Scale-Global-Opt' and shown to be the second-best strategy in the original paper. In this strategy only $(\epsilon, \beta)$ we optimised using 100 uniformly distributed samples on the intervals (0.5, 1.5) and (0.05, 0.15) respectively. As for all other models, we sample only one observation per parameter point. Unlike the rest of the comparison methods, KELFI has not been adapted yet to use active learning strategies. Therefore, we report results only for the 1000 total observations.

\subsection{DGPs and GPs as surrogate models}
\label{sec:4-gpvsdgp}

In this section, we use the LV-2GP architecture as DGPs, and compare them with vanilla GPs, as surrogates for BOLFI. The LV-2GP is the simplest model that combines benefits of multiple GPs and an inclusion of a LV layer. We expected DGPs to have advantages over GPs for multimodal cases, and hoped for them to have similar performance and data-efficiency in the rest of the cases. This is not obvious, since DGP, as a more flexible model, is expected to have a larger variance. However, the difference turned out to be negligible in practice.

\begin{figure*}
    \centering
    \includegraphics[width=\textwidth]{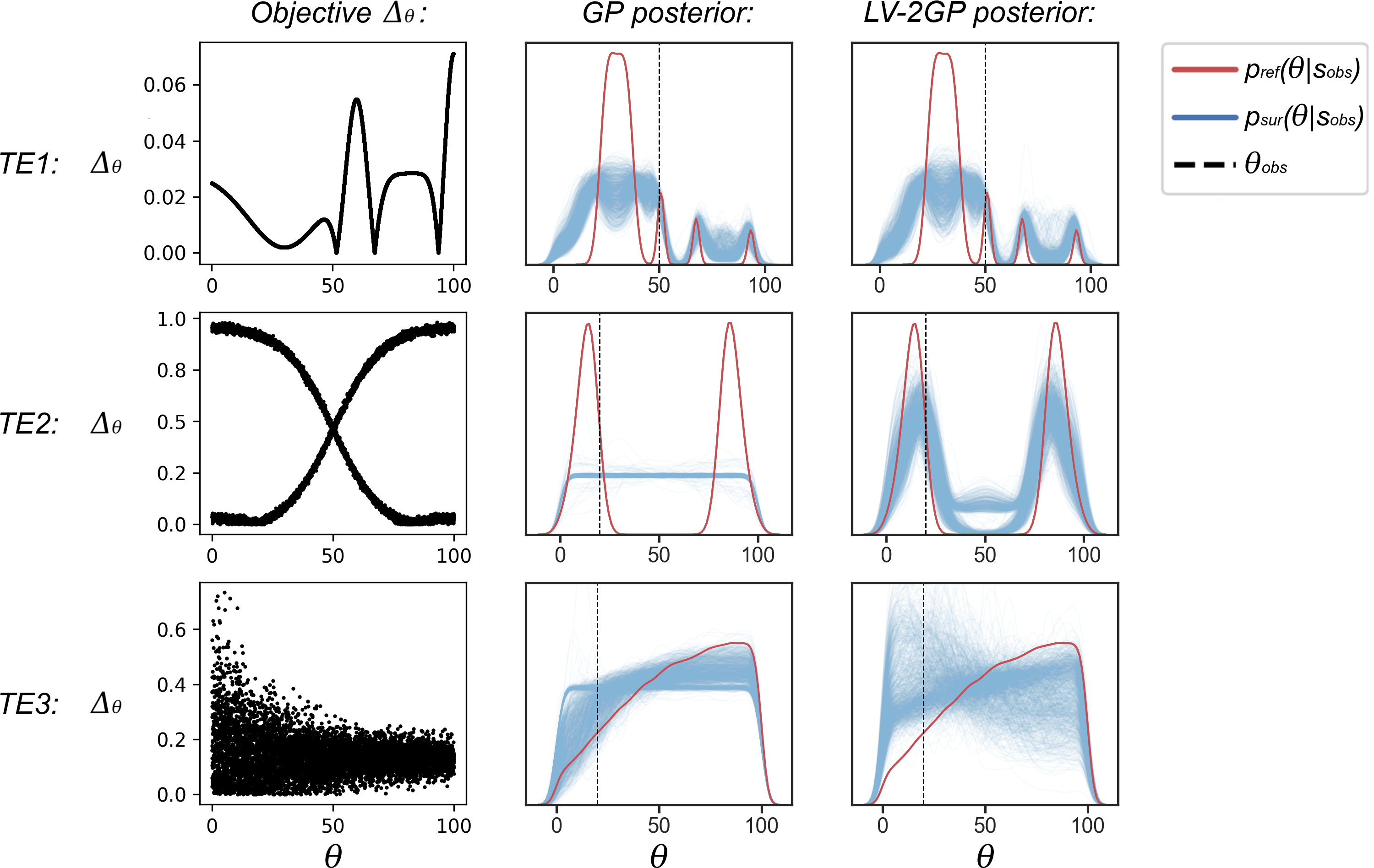}
    \caption{Approximation quality of posteriors by vanilla GP (middle column) and DGP (right column: LV-2GP is an instance of DGP), in three demonstration examples (rows). The figures show that DGPs maintain close approximation of the reference posterior (red lines) in \textbf{TE1} and \textbf{TE3}, and significantly surpass GPs in \textbf{TE2}. Both surrogates try to model the discrepancy functions (left column), and approximate the posterior (blue lines) of $\bt$ for $\dt \rightarrow 0$. The quality of inference can be inferred from how closely surrogate posteriors $p_\text{sur}(\bt | \sobs)$ with 1000 different initial evidence sets follow the reference posterior $p_\text{ref}(\bt | \sobs)$.}
    \label{fig:3-te}
\end{figure*}

\begin{figure*}
    \centering
    \subfloat[TE1]{\label{fig:2-te1}\includegraphics[height=3.8cm]{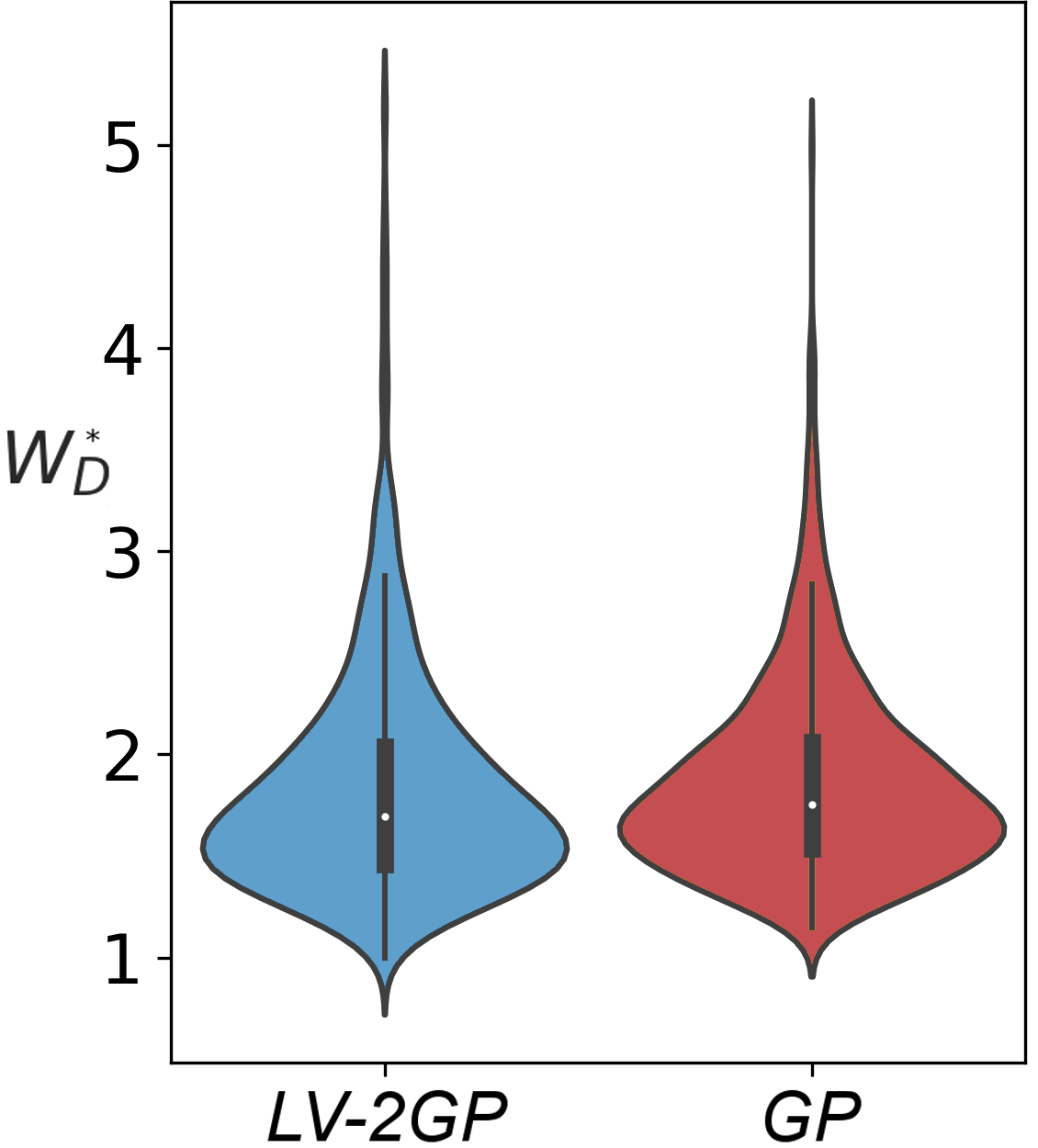}}
    \subfloat[TE2]{\label{fig:2-te2}\includegraphics[height=3.8cm]{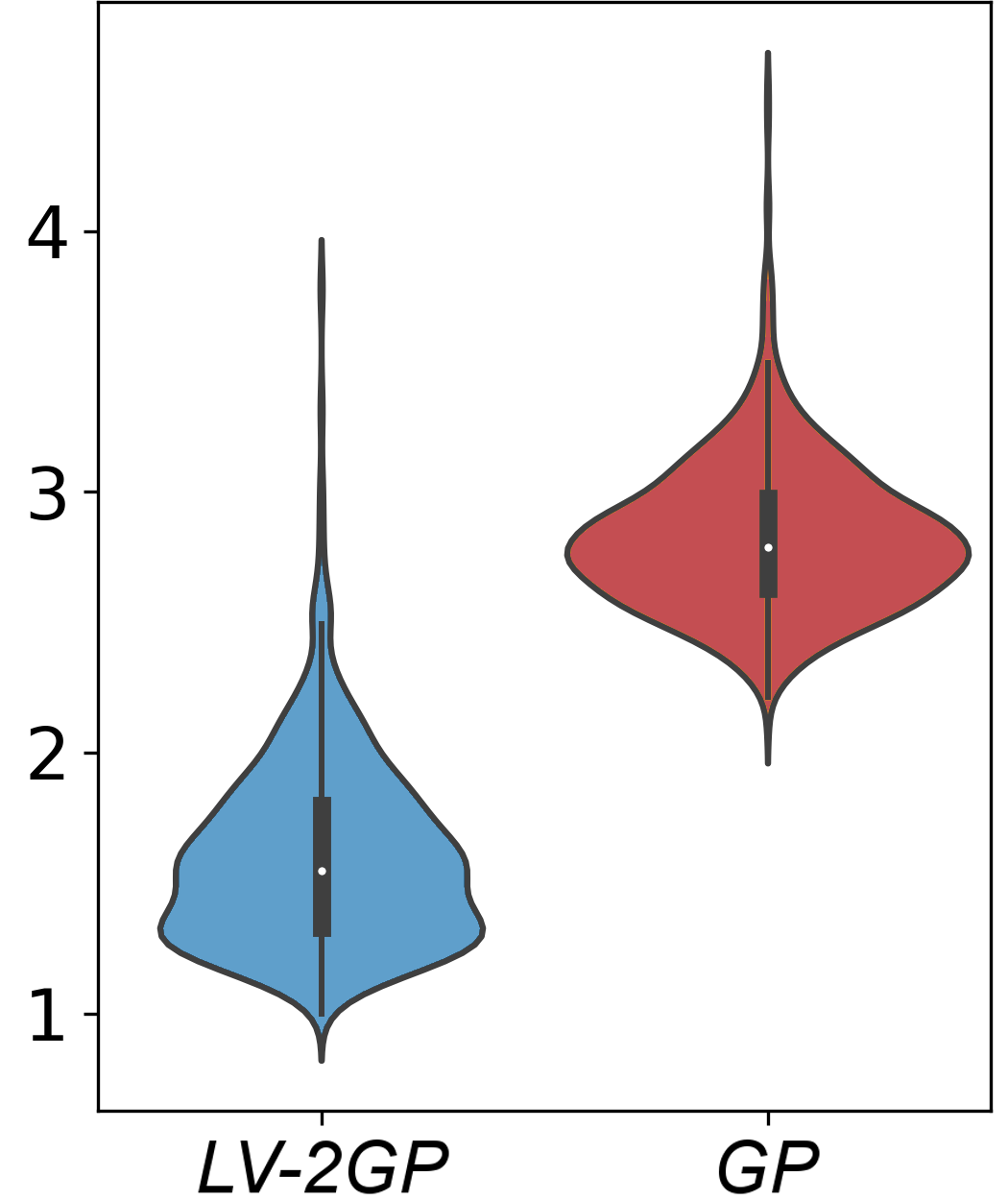}}
    \subfloat[TE3]{\label{fig:2-te3}\includegraphics[height=3.8cm]{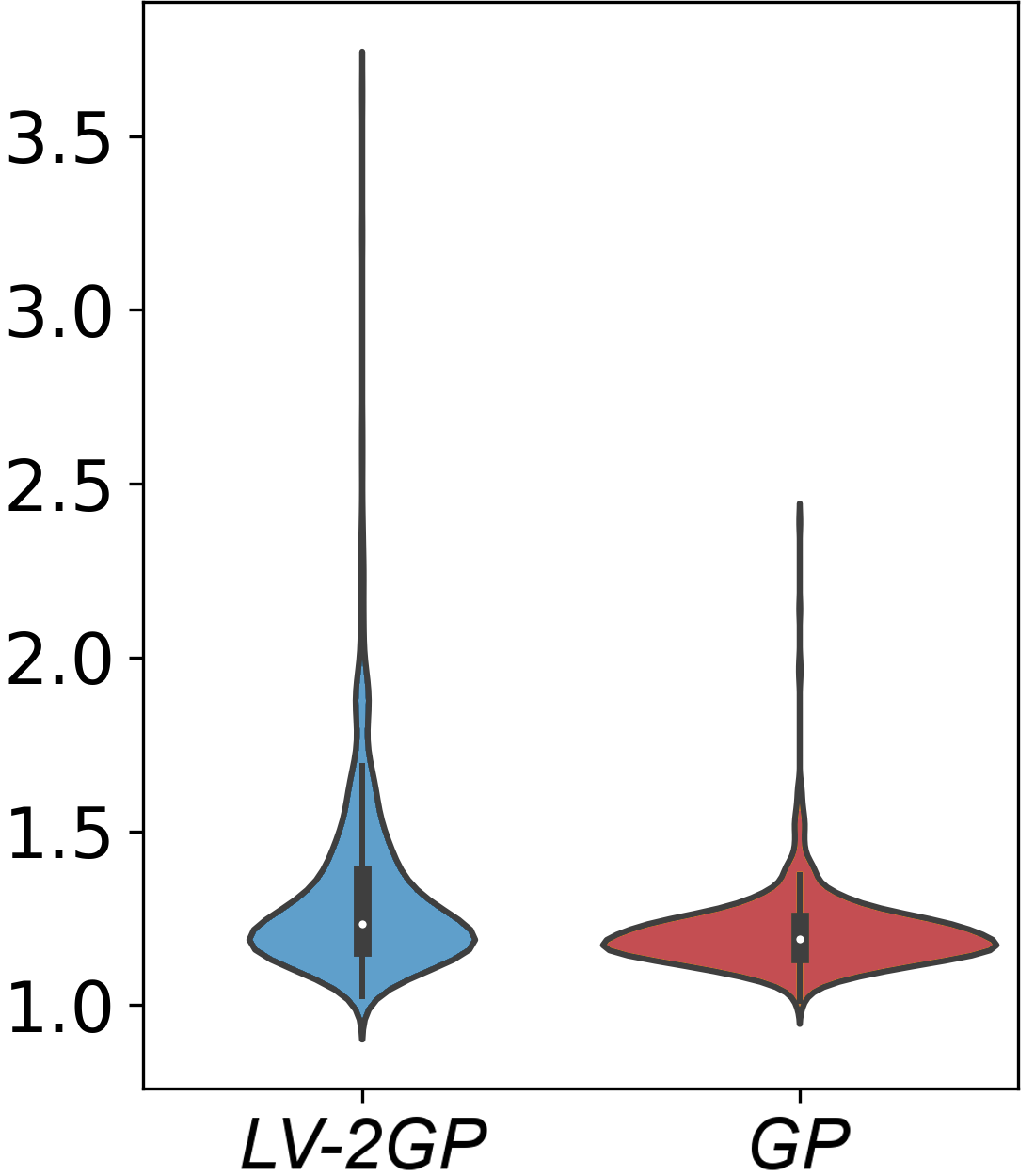}} 
    \subfloat[BDM]{\label{fig:2-bdm}\includegraphics[height=3.8cm]{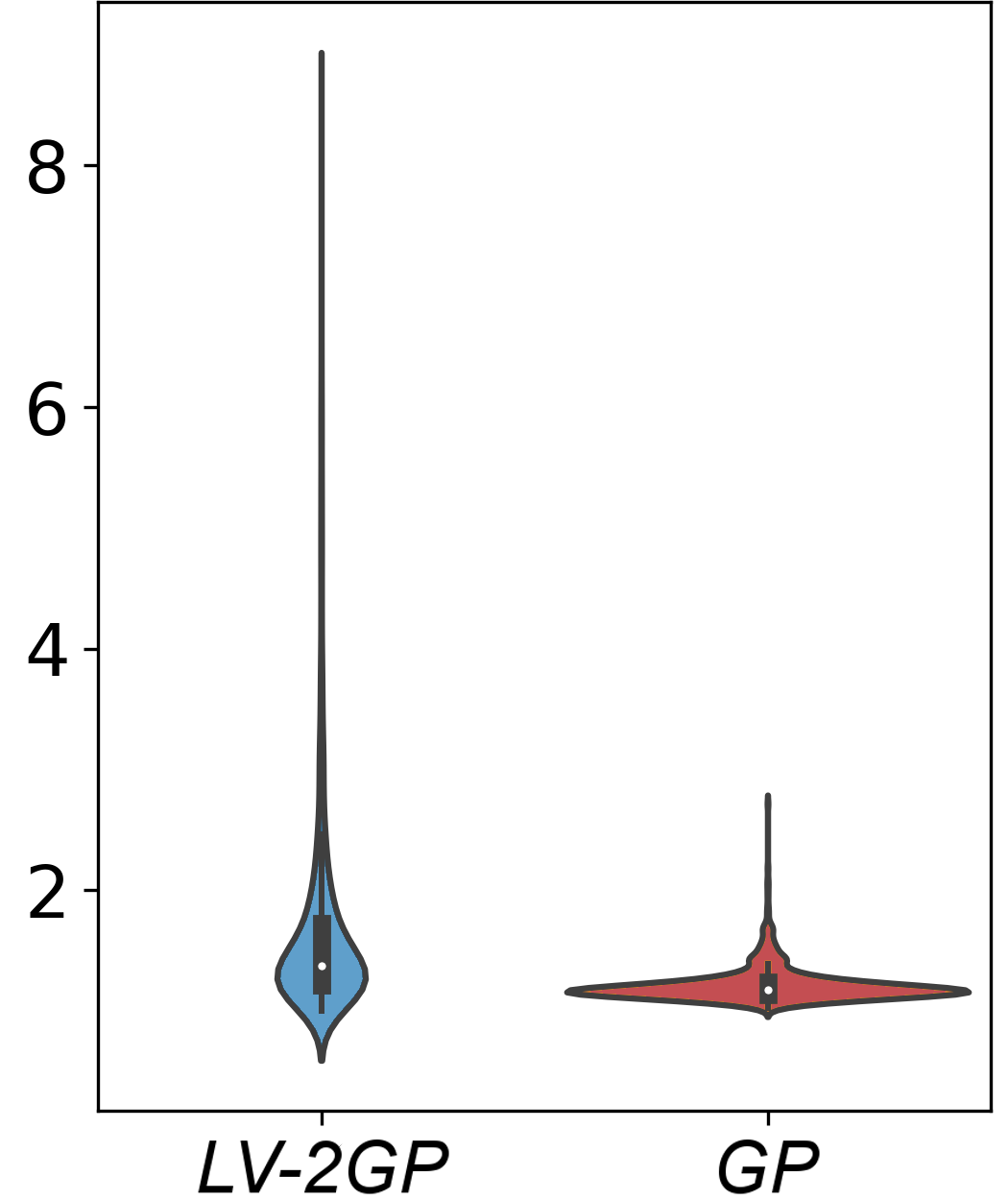}}
    \subfloat[NW]{\label{fig:2-nw}\includegraphics[height=3.8cm]{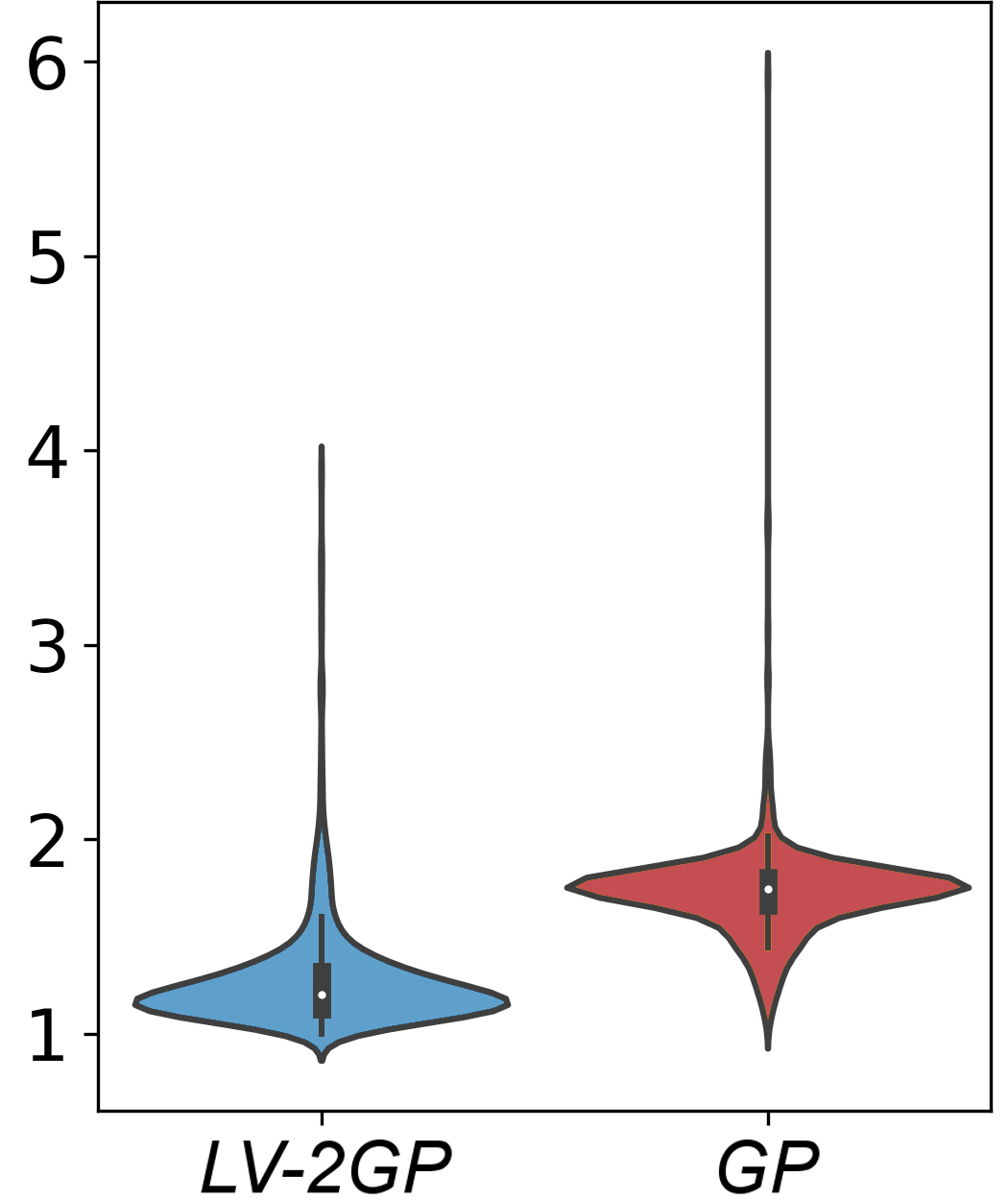}}
    \caption{Scaled Wasserstein distance between the surrogate models (GP and LV-2GP, an instance of DGP) posteriors and the true posterior of $\bt$; the smaller the distance, the better is the quality of approximations. The DGP approximations of the true posterior are better on multimodal \textbf{TE2} (b) and \textbf{NW} (e) examples, maintaining comparable performance on the rest of the cases. The white dot on the violin plot is the median, the black bar is the interquartile range, and lines stretched from the bar show lower/upper adjacent values.}
    \label{fig:2-wd}
\end{figure*}

We use one-dimensional toy examples (TE) to demonstrate the differences between GP and DGP surrogates on  three types of objective functions: non-stationary (TE1), multimodal (TE2) and heteroscedastic (TE3). Figure \ref{fig:3-te} clearly shows that DGPs can handle multimodality, while vanilla GPs cannot. In the TE2 case, the performance of both surrogates is very different, unlike the rest of the test cases. The DGP surrogate is able to capture the multimodal uncertainty, separating the two modes of the target distribution, while the GP surrogate fails to do so. DGPs and GPs perform similarly on non-stationary and heteroscedastic cases. In the non-stationary TE1, the posteriors look almost identical, while in TE3 the difference between posteriors is negligible due to a significant and complex noise component of the example, with DGPs having a higher variance. Such performance of both models on TE3 was expected, and demonstrates that DGPs, as a more flexible model, have greater flexibility than traditional GPs. The results of TE1 indicate that even though the DGP model can provide a better approximation of the whole function, it is sufficient to accurately represent the function at global minima. Overall, the toy example results strengthen our hypothesis about DGPs being able to handle objective functions with more irregular uncertainties, which is further confirmed with the scaled Wasserstein distances summarised in Figure  \ref{fig:2-te1}-\ref{fig:2-te3}.

\begin{figure*}
    \centering
    \subfloat[GP]{\label{fig:6-gp-nw}
    \includegraphics[height=3.3cm]{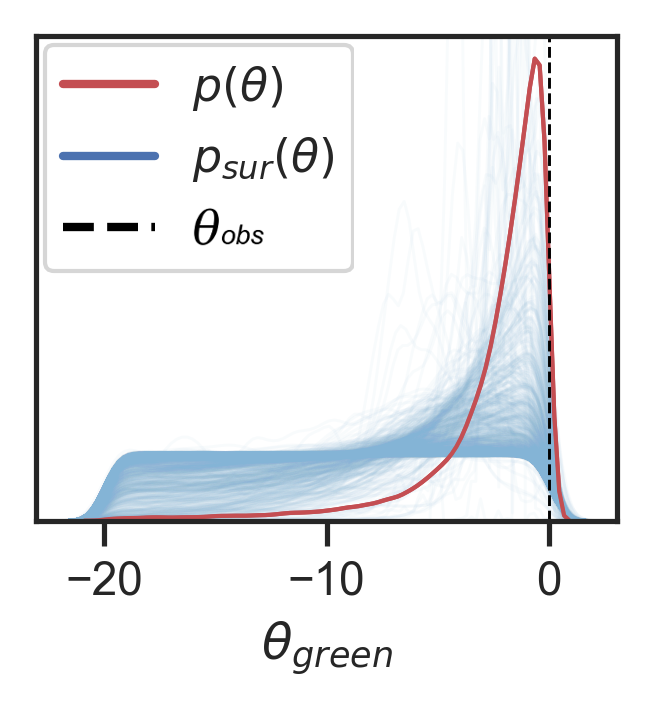}
    \includegraphics[height=3.3cm]{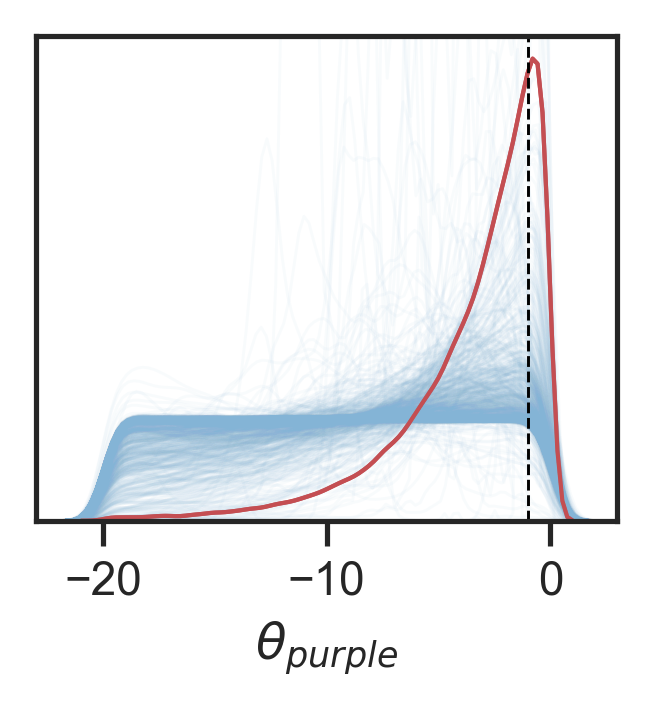}
    \includegraphics[height=3.3cm]{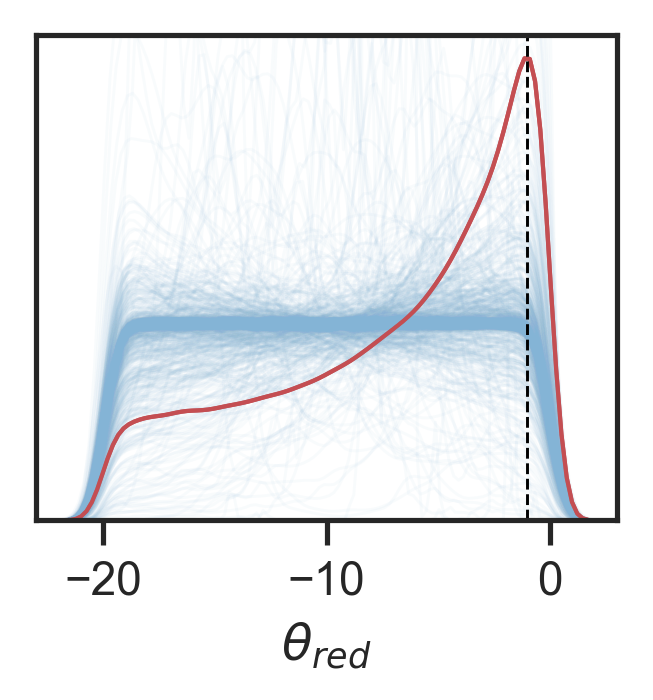}
    \includegraphics[height=3.3cm]{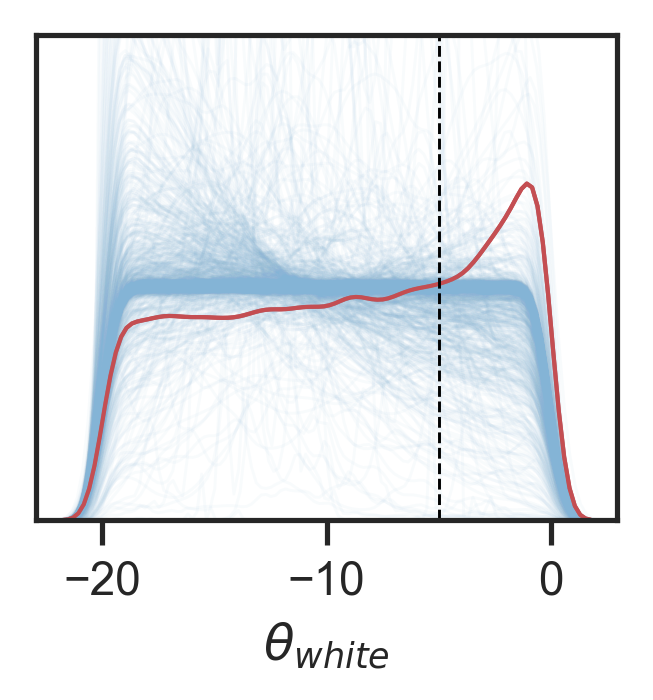}
    \includegraphics[height=3.3cm]{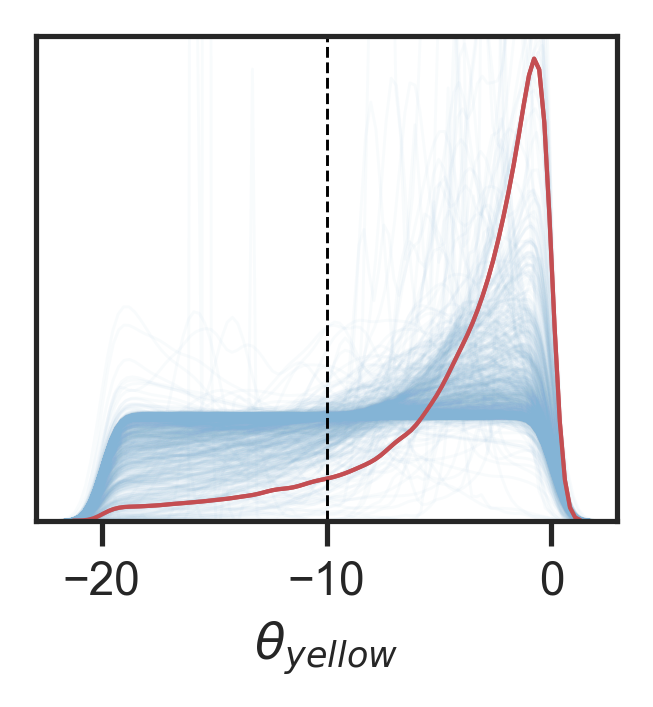}
    }
    \\
    \subfloat[DGP]{\label{fig:6-dgp-nw}
    \includegraphics[height=3.3cm]{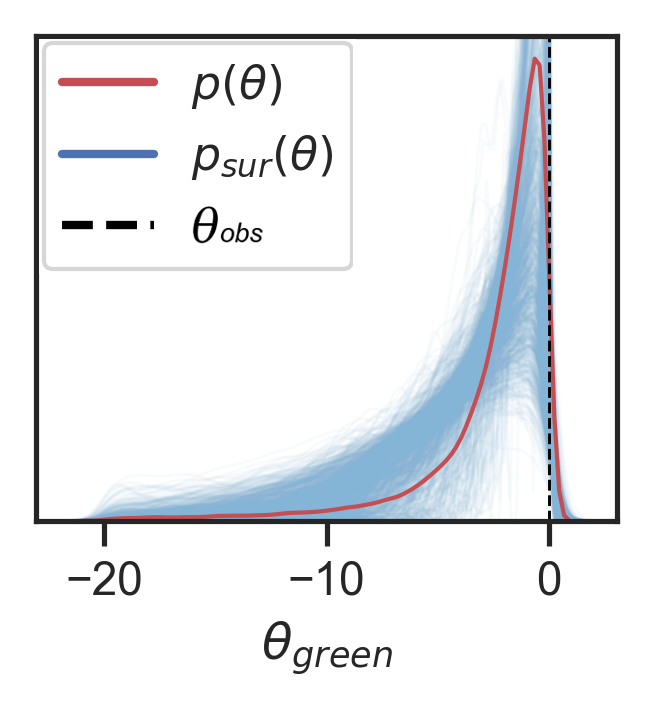} 
    \includegraphics[height=3.3cm]{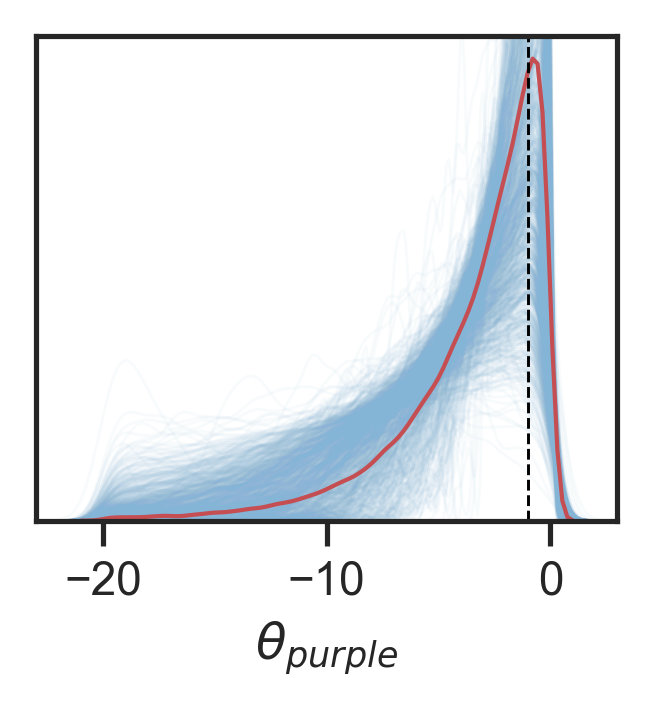}
    \includegraphics[height=3.3cm]{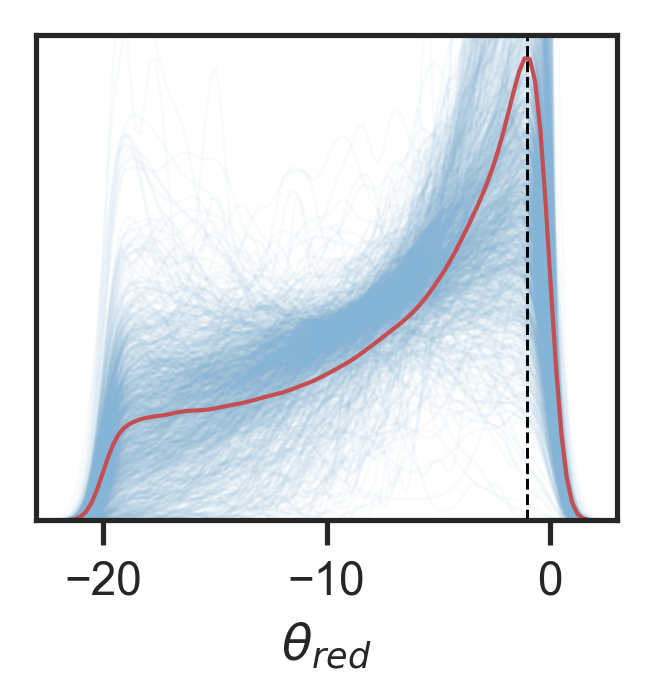}
    \includegraphics[height=3.3cm]{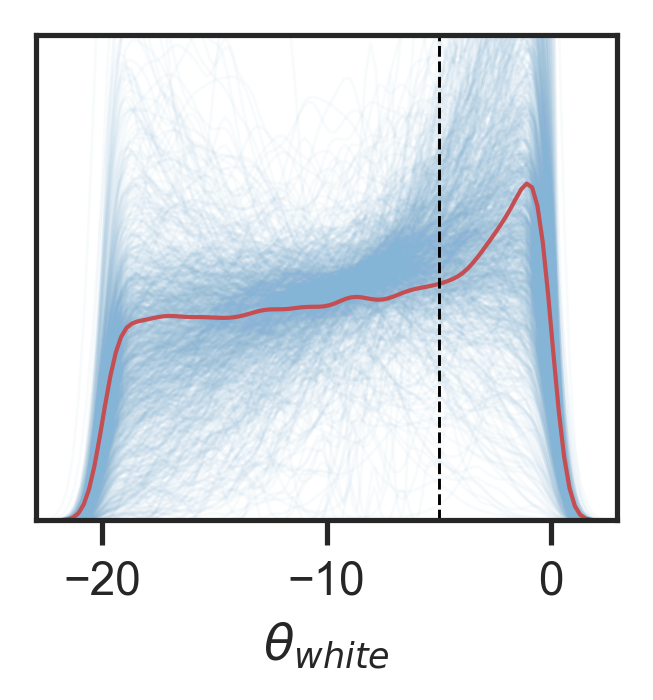}
    \includegraphics[height=3.3cm]{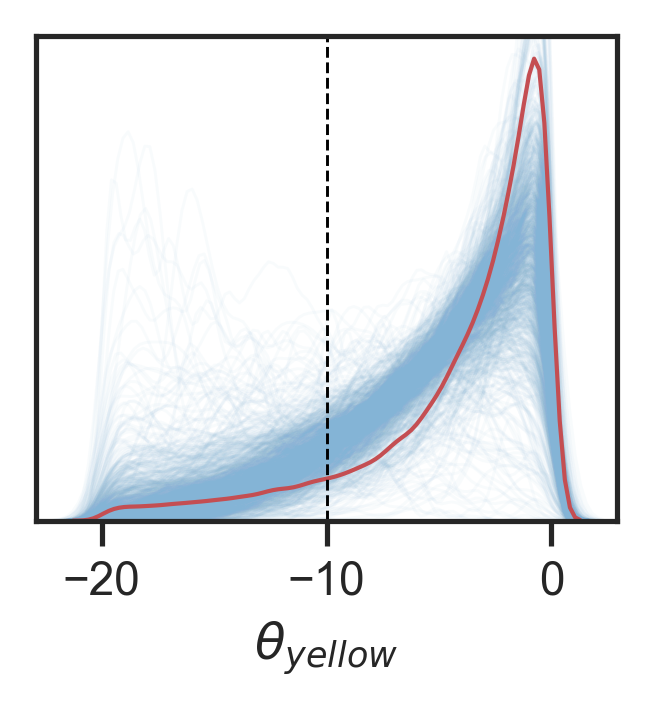}
    }
    \caption{Approximation quality of posterior marginals (columns) by vanilla GP (top) and DGP (bottom) in the \textbf{NW} case. Surrogate posterior samples (blue lines) of DGP provide a far more accurate approximation of the posterior marginals (red lines) than GPs.}
    \label{fig:6-nw}
\end{figure*}

\begin{figure*}
    \centering
    \subfloat[GP]{\label{fig:4-gp-bdm}
    \includegraphics[height=4.1cm]{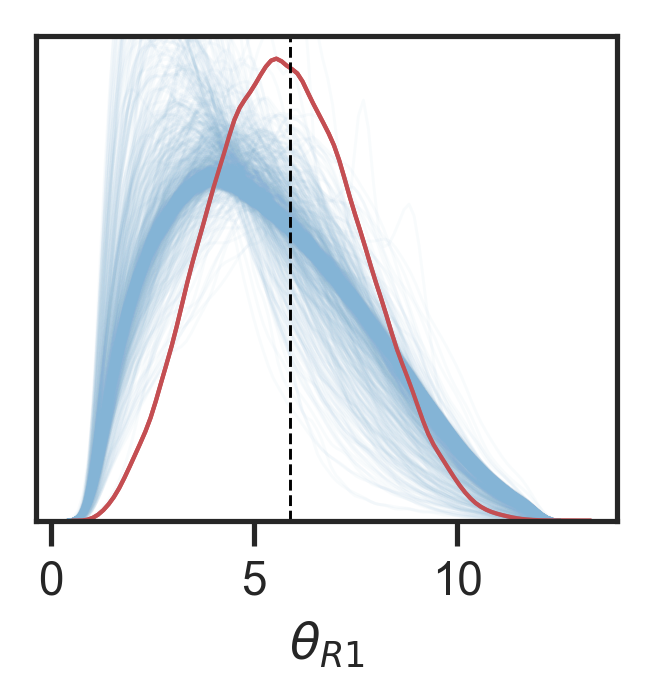}
    \includegraphics[height=4.1cm]{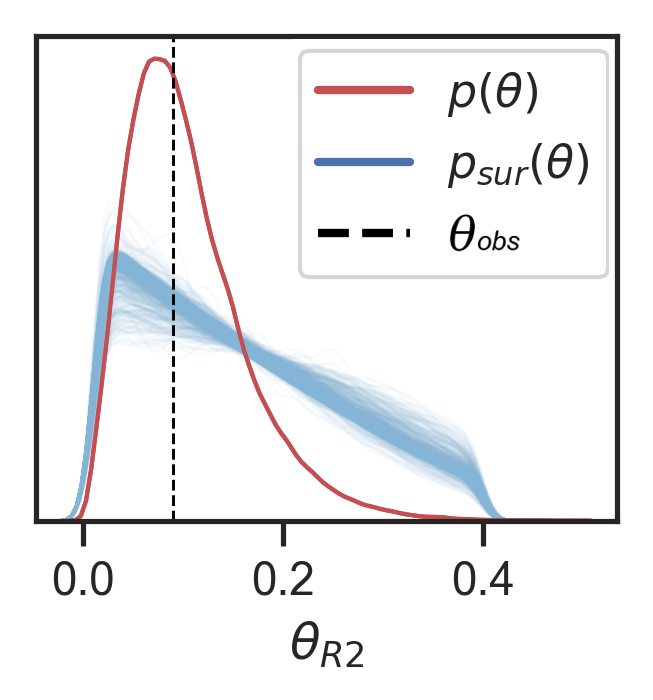}
    \includegraphics[height=4.1cm]{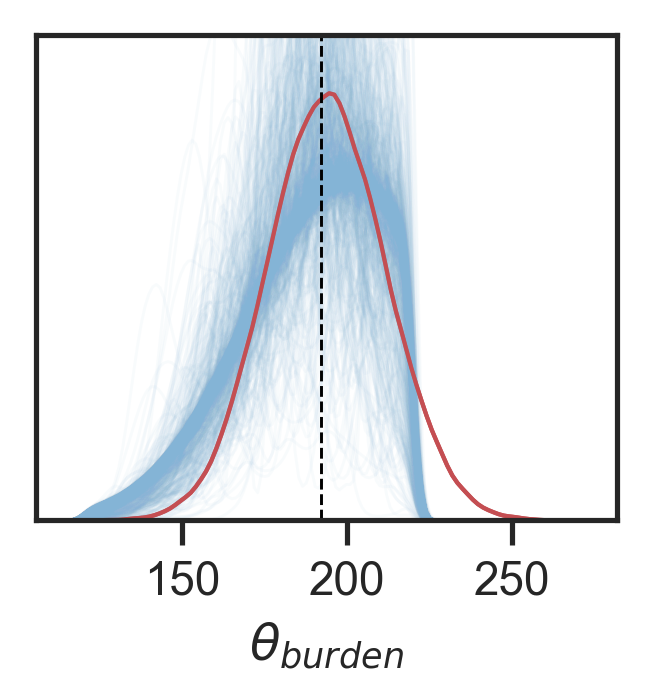}
    \includegraphics[height=4.1cm]{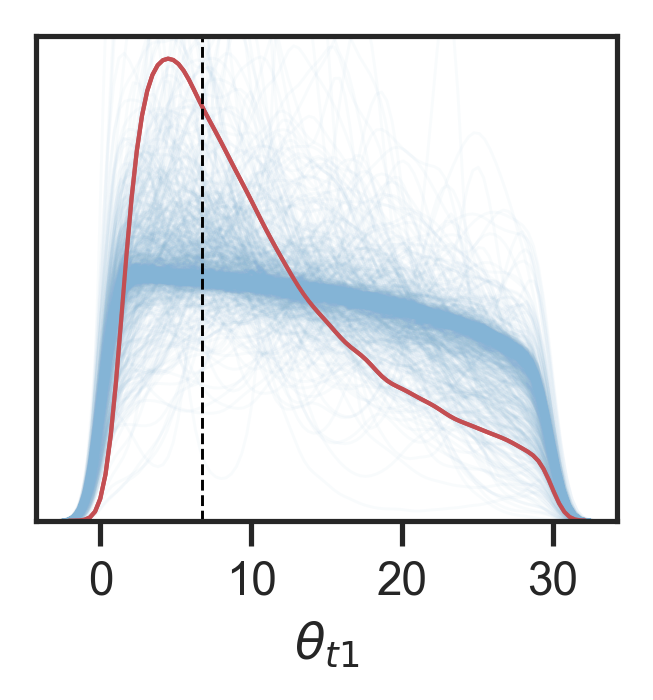}
    }
    \\
    \subfloat[DGP]{\label{fig:4-dgp-bdm}
    \includegraphics[height=4.1cm]{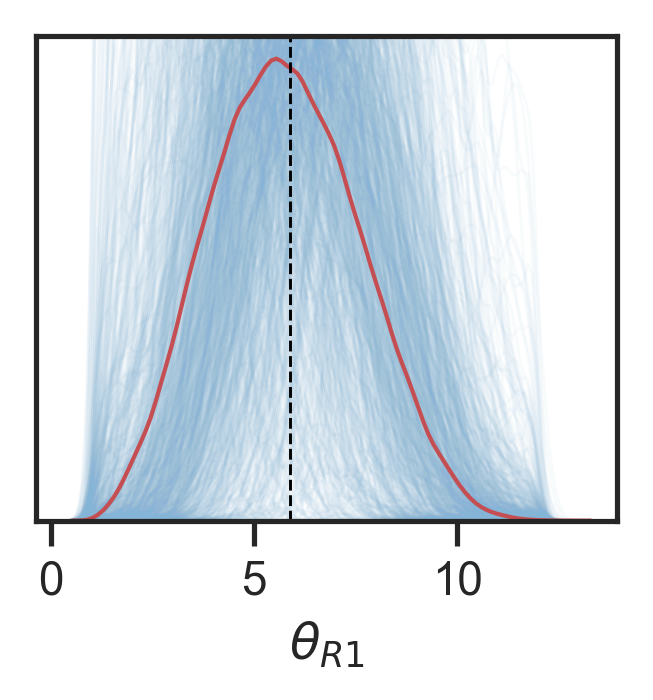} 
    \includegraphics[height=4.1cm]{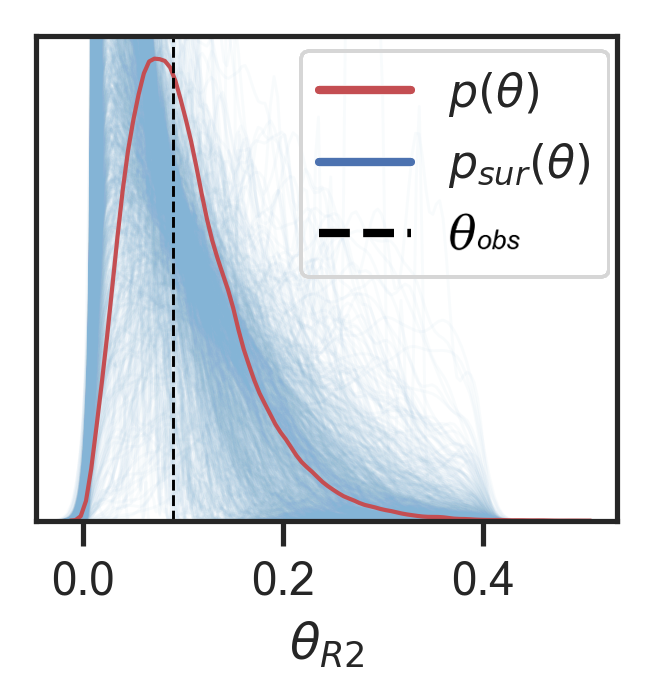}
    \includegraphics[height=4.1cm]{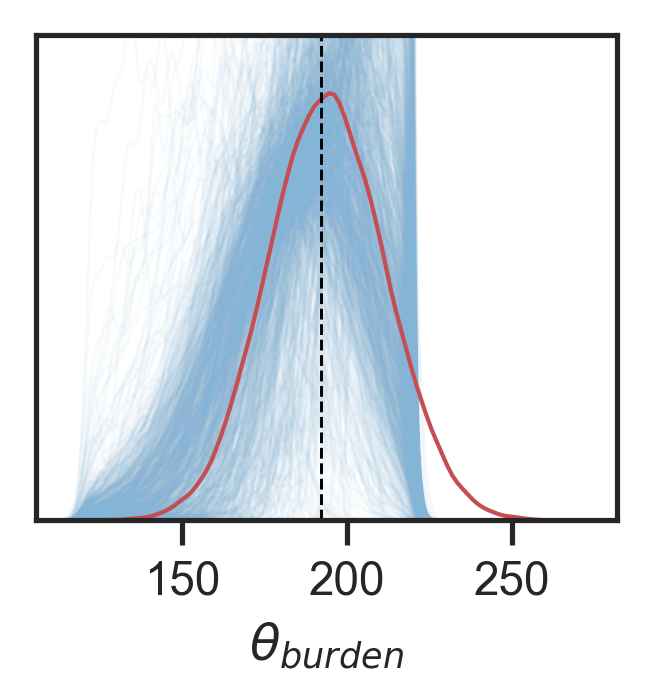}
    \includegraphics[height=4.1cm]{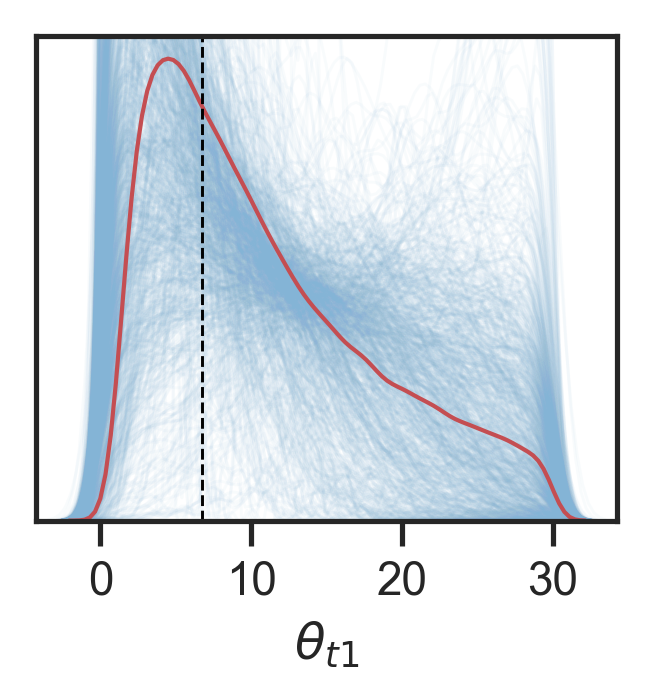}
    }
    \caption{Approximation quality of posterior marginals (columns) by vanilla GP (top) and DGP (bottom) in the \textbf{BDM} case. Surrogate posterior samples (blue lines) of DGPs are flexible enough to approximate all marginals of the true parameter $\bt$ posterior (red lines), though have higher variance, while GP samples converge to poor marginal distributions.}
    \label{fig:4jp-BDM}
\end{figure*}

In the studies on BDM and NW, DGPs are either better or on the same level as vanilla GPs in approximating the posterior. The results for the NW case from Figure \ref{fig:2-nw} clearly separate the performance of DGPs, suggesting that they successfully model multimodality in higher dimensions. Additionally, DGP samples (Figure \ref{fig:6-dgp-nw}) approximate marginals much more accurately than GP samples (Figure \ref{fig:6-gp-nw}), which also suggests an improvement over GPs in a multimodal planning environment. On the other hand, the BDM performance of DGPs is comparable to GPs, but offers no clear advantage over GPs because of higher variance, as shown in \ref{fig:2-bdm}. A closer inspection of individual marginals reveals that DGP marginal posterior samples (Figure \ref{fig:4-dgp-bdm}) are flexible enough to approximate true distributions at least for $\theta_{R1}, \theta_{R2}$ and $\theta_{t1}$, while GP samples (Figure \ref{fig:4-gp-bdm}) seem to converge to wrong marginals. In summary, DGPs unlike GPs can work with both multimodal and unimodal uncertainties, making them especially suitable for cases when no prior information about the form of the uncertainty is available.

\begin{figure*}[tb]
    \centering
    \subfloat[NW]{\label{fig:5-de-NW}\includegraphics[width=.49\textwidth]{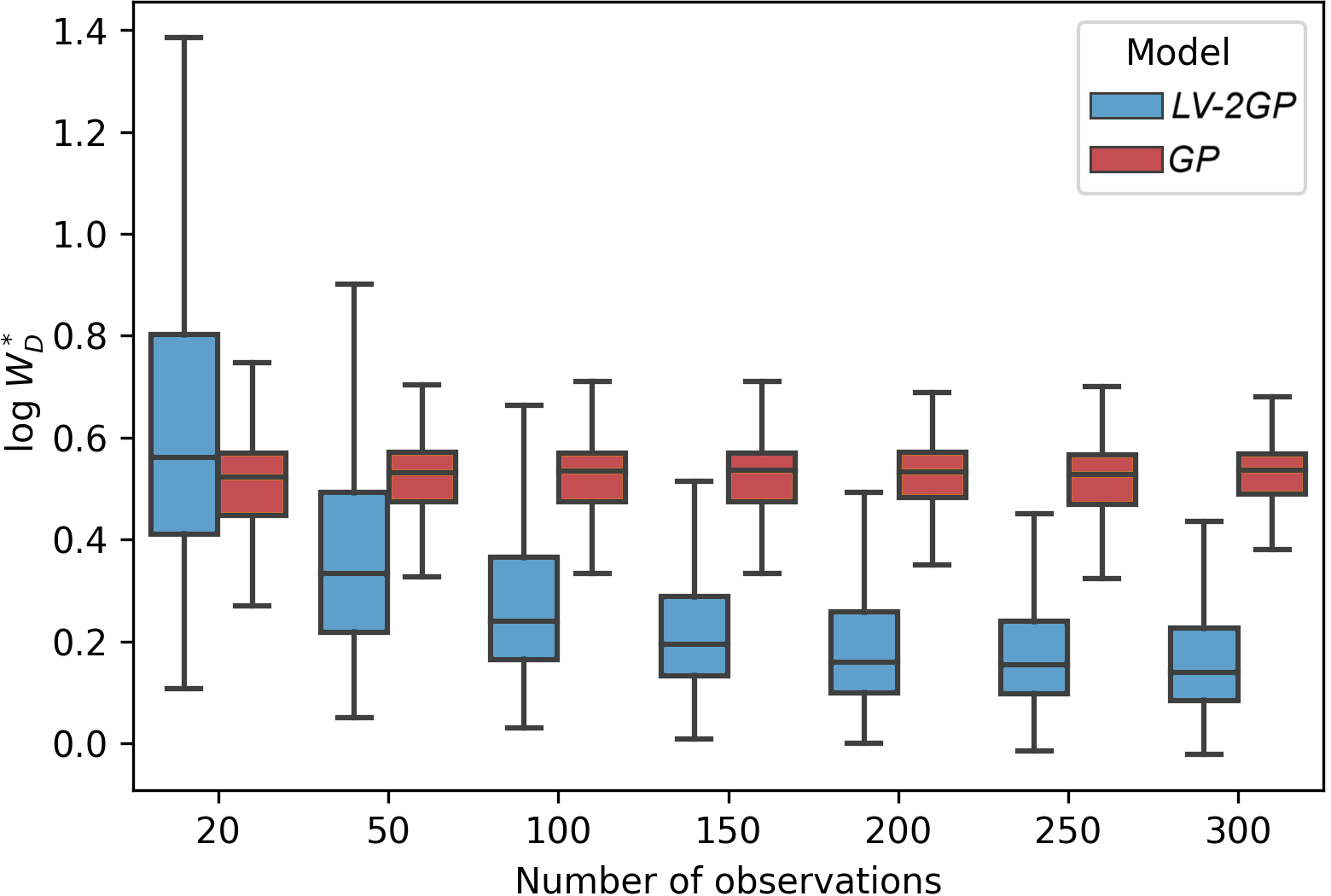}}
    \hfill
    \subfloat[BDM]{\label{fig:5-de-BDM}\includegraphics[width=.49\textwidth]{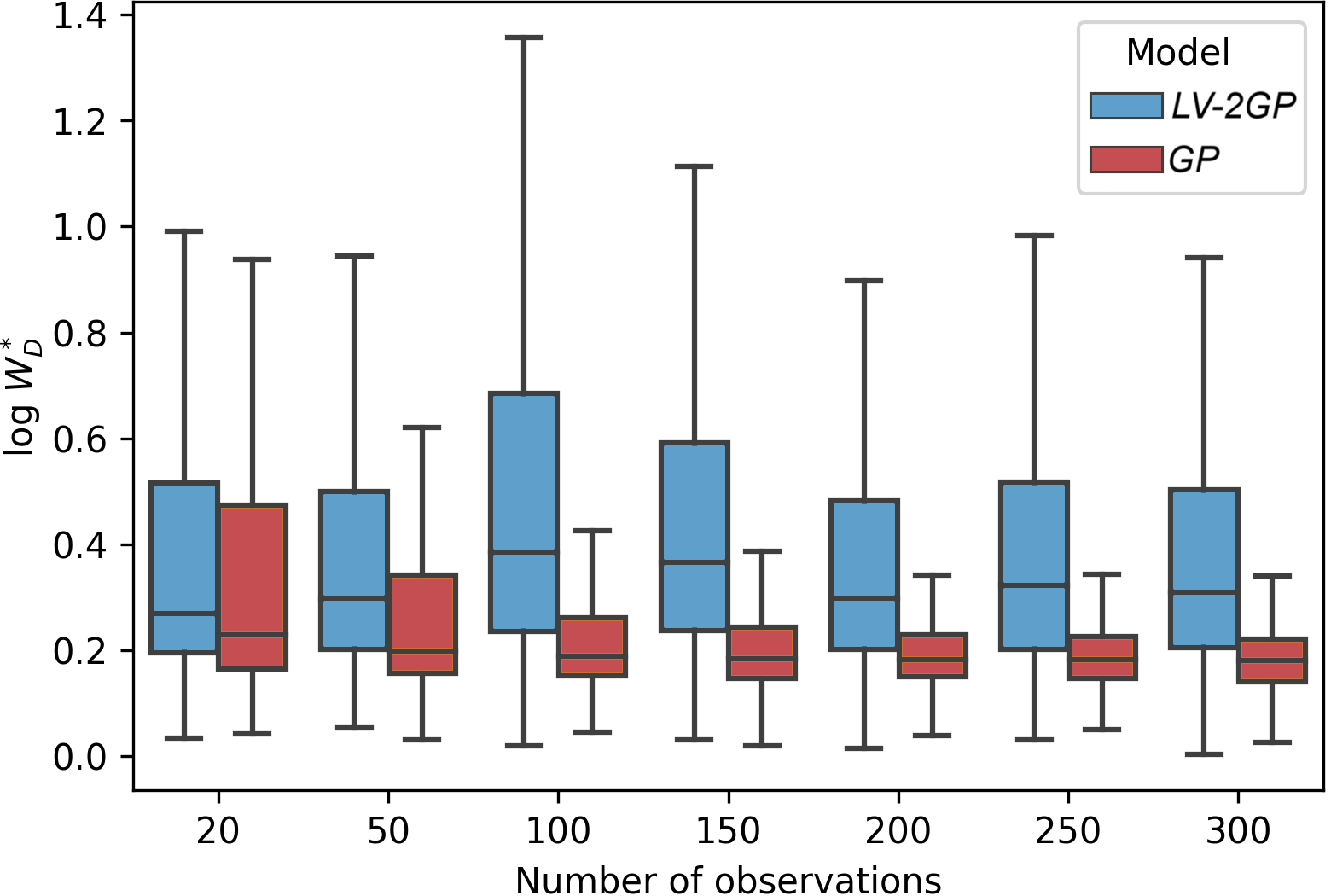}}
    \caption{Log-scaled Wasserstein distance between surrogate posteriors and the true posterior, shown in the \textbf{NW} and \textbf{BDM} experiments, as a function of the number of total observations. DGP approximation accuracy scales better in the \textbf{NW}, and demonstrates similar data-efficiency as GPs in the \textbf{BDM}. The box plots were computed with distances across 1000 simulations. The horizontal line on box plots shows the median, the bar shows upper and lower quartiles, and the whiskers indicate the rest of the quartiles.}
    \label{fig:de}
\end{figure*}

We conducted additional experiments to evaluate the performance of DGPs and GPs under different simulation budgets in the considered case studies. In the NW case (Figure \ref{fig:5-de-NW}), DGP steadily improves performance with more observations, clearly outperforming GP with more than some tens of observations. In contrast, the scaling behaviour of both methods in the BDM case (Figure \ref{fig:5-de-BDM}) is similar, but GP is somewhat better, and in particular has a lower variance for large numbers of observations. The explanation of the difference between the cases is that NW is clearly multimodal, whereas BDM is not, at least visibly. Hence, standard GP is sufficient for modelling BDM, resulting in lower variance, but not for modelling NW. In conclusion, DGPs can perform very similarly to GPs on the small-data tasks, and being a more flexible model, can further improve the approximations with more data. These results support our claim of DGPs being capable of modelling multimodal target distributions with a limited number of function evaluations.

\subsection{Comparison of LFI approaches}
\label{sec:4-addexp}

The comparison of the proposed DGP surrogates to other LFI approaches (Table \ref{tab:wd}) show that DGPs outperform MAF, MDNs and KELFI alternatives. None of those methods achieve a performance comparable to DGPs, even with much more data (1000 observations vs 200). Even though MAF and MDNs use active learning, they are trying to model the likelihood directly, in contrast to DGPs that model the discrepancy. The former is a more general and harder problem, that requires many more observations with the benefit of not having to retrain the model if the observed data is changed. On the other hand, KELFI does not use any active learning strategies, therefore, it was expected to have worse data-efficiency than DGPs. The only exception, where those alternative techniques performed slightly better than DGPs, is the MDNs model in the BDM case. This can be explained by the Gaussian form of the BDM posterior, where MDNs have an advantage, since they model Gaussian mixtures. In summary, all the considered alternatives have the necessary flexibility to show good performance on the considered cases, however, they require significantly more data than DGPs, making them unsuitable for modelling irregularly behaved distributions in a small data setting. Therefore, DGPs is the preferable candidate for doing LFI with computationally expensive simulators.

We compared different architectures of the LV-DGP model, assessing the influence of the number of the LV and GP layers. Results in Table \ref{tab:wd} show that the inclusion of the LV layer significantly affected the performance of the considered models. The LV layers both increased the performance of the models on multimodal TE2 and NW examples, and slightly increased the variance in all other cases. This is expected, since architectures without the LV layer can be considered more restricted and their predictive posterior distributions will have less variance across runs. More observations are likely to reduce the variance of the architectures with the LV layer and improve the accuracy further. As for GP layers, deeper architectures did not show significantly better results. The considered small data setting, which is implied by computationally expensive simulators, does not benefit from having multiple GP layers. In summary, 'LV-2GP' can be considered as a preferred architecture, since it is the simplest model that has most of the benefits from the inclusion of the LV layer.

\section{Discussion}

We introduced a novel method for statistical inference when the likelihood is not available, but drawing samples from a simulator is possible, although computationally intensive. The introduced method is an extension of BOLFI \citep{gutmann2016bayesian} where we have adopted DGP surrogates instead of GP surrogates to model the relationship of the parameters and the stochastic discrepancy between observed data and simulated data. These new surrogates use quantile-based modifications for an acquisition function and likelihood approximation, making it feasible for LFI problems. The proposed extension retains the active learning property of BOLFI so that the posterior distribution is sought out with as few samples as possible. The flexibility of the DGPs improved the resulting posterior approximations in cases where flexibility was required, and otherwise the observed performance was similar in both cases. Especially good improvements were observed in cases where the distribution of the discrepancy was multimodal, i.e.~in cases where GP is known to perform poorly as an estimator.

The improvements from using DGP surrogates come with increased computational cost, which we demonstrated to be negligible for computationally heavy simulators. DGPs also had a higher variance in an unimodal higher dimensional example. Even though data-efficiency experiments indicated that the DGP variance improves with more observations, a major contribution to this high variance is likely related to the ability to model multimodality. Comparison methods, that showed this ability as well, had similar variance in the unimodal case. The best neural density and kernel mean embedding based alternatives were outperformed by DGPs in a grand majority of cases, providing better approximations with fewer available data. We recommend using DGPs in cases with complicated target distributions, where their more expressive surrogates are needed and work better than vanilla GPs.

A natural progression of this work is to analyse DGP uncertainty decomposition and its propagation through layers. Decomposing the uncertainty into its aleatoric and epistemic components would allow better exploration of the parameter space. This is especially important when dealing with multimodal distributions, since they often have high epistemic uncertainty that may prevent BO from exploring other parameter regions. As for uncertainty propagation, individual layers of DGPs can be used to learn intermediate transitions inside the simulator. Doing so would require opening the black-box of the simulator and incorporating these intermediate transitions as data in the training process. This additional information would lead to better usage of simulator time, since some futile simulations could be abandoned once their transition variables become available. In conclusion, better uncertainty decomposition and propagation can further improve data-efficiency of LFI, when dealing with computationally expensive simulators that have irregular noise models.

\section{Declarations}

\subsection{Funding}
This work was supported by the Academy of Finland (Flagship programme: Finnish Center for Artificial Intelligence FCAI; grants 328400, 325572, 319264, 292334). Authors HP and JC were also supported by European Research Council grant 742158 (SCARABEE, Scalable inference algorithms for Bayesian evolutionary epidemiology). Computational resources were provided by the Aalto Science-IT Project.

\subsection{Declarations of interest}
None.

\subsection{Availability of data and material}
Not applicable.

\subsection{Code availability}
All code is available through the link below:

\href{https://github.com/alexaushev/LFIwithDGPs}{https://github.com/alexaushev/LFIwithDGPs}

\bibliographystyle{unsrtnat}
\bibliography{main}   

\appendix

\section{LV-DGP inference in multimodal cases}
\label{sm:iwvi-sghmc}

We considered an alternative inference method for LV-DGPs, namely Stochastic Gradient Hamiltonian Monte Carlo (SGHMC)  \citep{havasi2018inference}. The inference uses Moving Window Markov Chain Expectation Maximisation method for the hyperparameter maximisation, where a set of samples is maintained throughout the inference and random samples are used to calculate the gradient for hyperparameters optimisations. We used a window of 100 samples with 20,000 Adam optimisation steps with  learning rate $5 \cdot 10^{-3}$ and learning rate decay of 0.99 (optimisation parameters are the same as in \citet{salimbeni2019deep}). The burn-in phase consisted of 20,000 iterations, followed by the sampling phase over the course of 10,000 iterations. For convergence diagnostics, we retained every 50th sample with the set size of 1000 samples for each one of four chains. The R-hats \citep{gelman1992inference} are reported for each of the 50 inducing points that are sampled with the SGHMC. This inference technique was compared against Importance-Weighted Variational Inference (IWVI) by \citet{salimbeni2019deep} for the 'LV-2GP' architecture in the TE2 and NW cases.

\begin{figure*}
    \centering
    \subfloat[TE2]{\label{fig:sm-te2}\includegraphics[height=5.4cm]{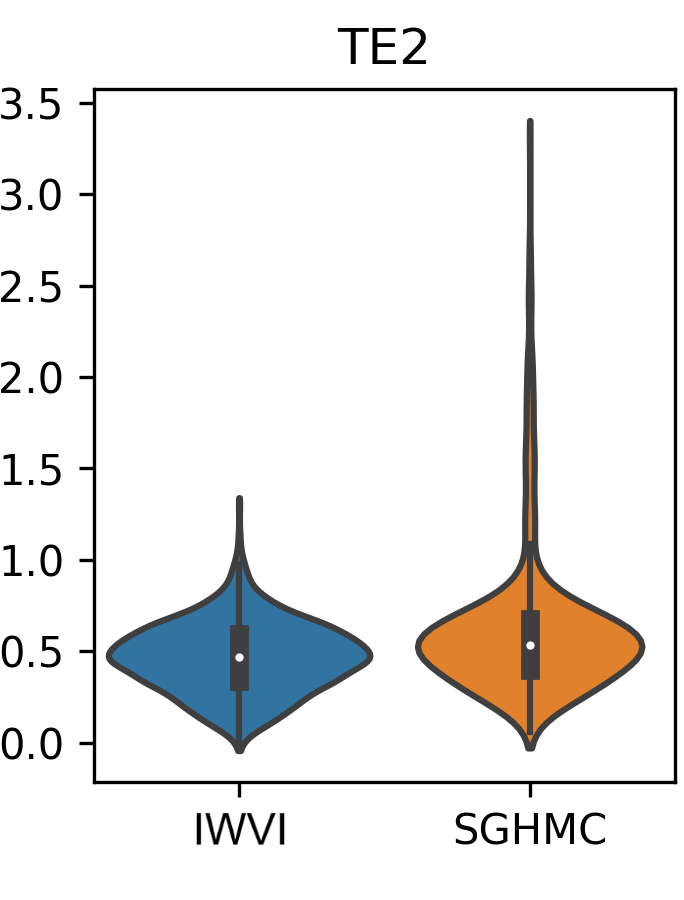}}
    \hfill
    \subfloat[NW]{\label{fig:sm-nw}\includegraphics[height=5.4cm]{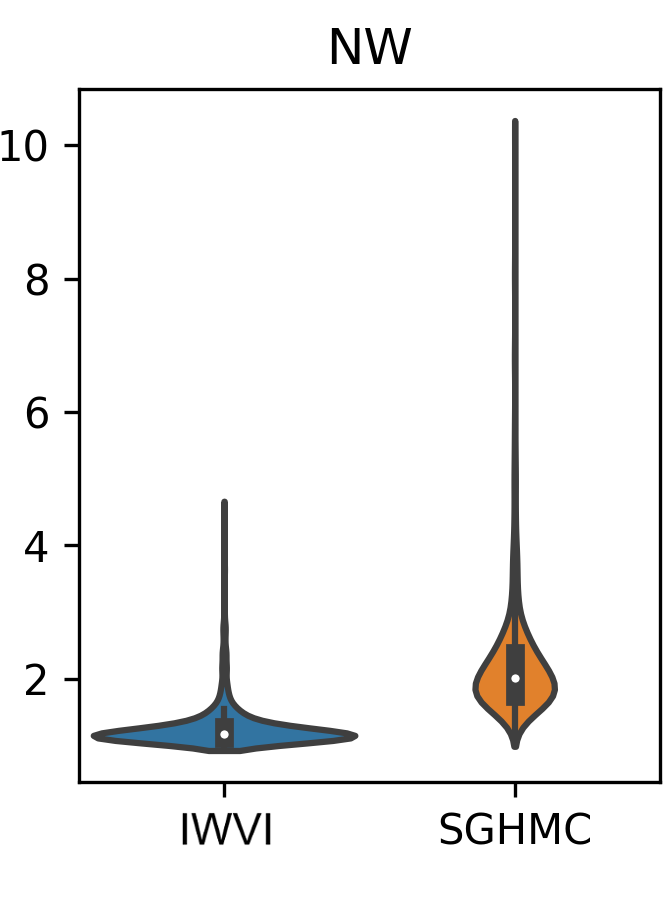}}
    \hfill
    \subfloat[SGHMC convergence diagnostics]{\label{fig:sm-rhat}\includegraphics[height=5.4cm]{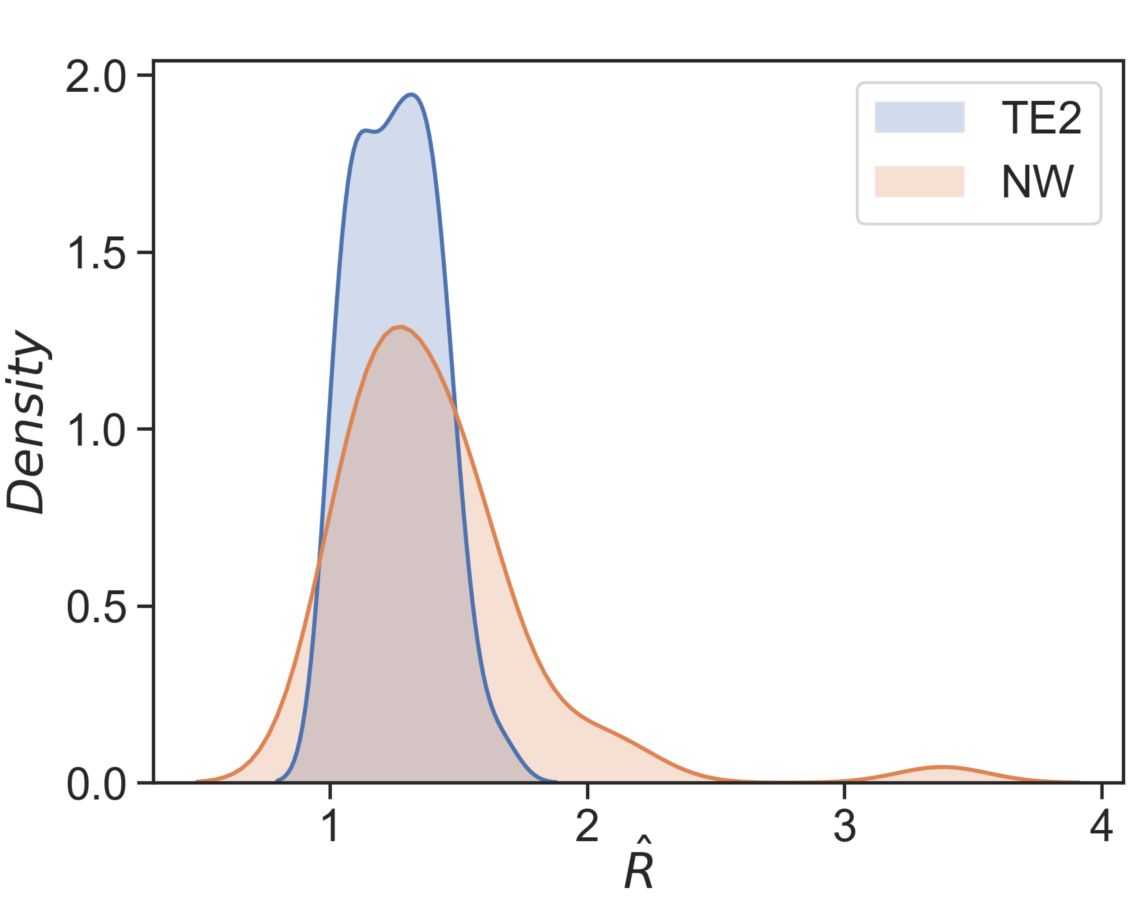}}
    \caption{The IWVI posterior is as close to the true posterior as the SGHMC posterior for the LV-2GP architecture in one-dimensional \textbf{TE2} (a) and much further away from it in the high-dimensional \textbf{NW} (b). The convergence diagnostics $\hat{R}$ (c) of 50 DGP inducing outputs show that the SGHMC has likely failed to converge on the higher-dimensional case, since most of the density mass is far beyond the acceptable $\hat{R}$ values of around $1.1$. The scaled Wasserstein distance between the IWVI and SGHMC posteriors and the true posterior of $\bt$ is used for the performance comparison in \textbf{TE2} (a) and \textbf{NW} (b); the smaller the distance, the better is the quality of approximation. The white dot on the violin plot is the median, the black bar is the interquartile range, and lines stretched from the bar show lower/upper adjacent values.}
    \label{fig:iwvivssghmc}
\end{figure*}

The performance of IWVI and SGHMC techniques for the same 'LV-2GP' architecture suggests that the variational inference posterior $q(u)$ follows the SGHMC approximation, at least for the one-dimensional multimodal case. Both inference methods showed almost exactly the same performance for the TE2 example, with SGHMC having slightly higher variance, as shown in Figure \ref{fig:sm-te2}. In higher dimensions, SGHMC performed much worse, see Figure \ref{fig:sm-nw}. This suggests that IWVI posterior followed SGHMC approximation closely in the TE2 example, but for the NW case results are inconclusive, since SGHMC likely failed to converge (Figure \ref{fig:sm-rhat}. In summary, LV-DGP with IWVI seems to be the preferable solution for doing LFI in the small data setting.

\end{document}